\pdfoutput=1

\documentclass[11pt]{article}

\usepackage[]{acl}

\usepackage{times}
\usepackage{latexsym}

\usepackage[T1]{fontenc}

\usepackage[utf8]{inputenc}
\usepackage{float}
\usepackage{microtype}

\usepackage{inconsolata}
\usepackage{graphicx}
\usepackage{subfig}
\usepackage{multirow}
\usepackage{amsfonts}
\usepackage{booktabs}
\usepackage{stfloats}
\usepackage{amsmath}
\usepackage{amssymb}
\usepackage{xcolor}
\definecolor{deepgreen}{RGB}{17,139,53}
\definecolor{deepred}{RGB}{231,82,122}
\usepackage[ruled,linesnumbered]{algorithm2e}
\title{Towards Objective Fine-tuning:\\How LLMs' Prior Knowledge Causes Potential Poor Calibration?}

\author{
Ziming Wang\textsuperscript{1}\thanks{\ \ Equal contribution.}, 
Zeyu Shi\textsuperscript{1}\footnotemark[1], 
Haoyi Zhou\textsuperscript{2,3}\thanks{\ \ Corresponding author.}, 
Shiqi Gao\textsuperscript{1}, \\
\textbf{Qingyun Sun}\textsuperscript{1}, 
\textbf{Jianxin Li}\textsuperscript{1,3} \\
\textsuperscript{1}SKLCCSE, School of Computer Science and Engineering, Beihang University \\
\textsuperscript{2}School of Software, Beihang University \\
\textsuperscript{3}Zhongguancun Laboratory, Beijing \\
\texttt{\{wangzm412, szy\_629, haoyi, gaoshiqi, sunqy, lijx\}@buaa.edu.cn}
}

\begin{document}
\maketitle
\begin{abstract}
Fine-tuned Large Language Models (LLMs) often demonstrate poor calibration, with their confidence scores misaligned with actual performance. While calibration has been extensively studied in models trained from scratch, the impact of LLMs' prior knowledge on calibration during fine-tuning remains understudied. Our research reveals that LLMs' prior knowledge causes potential poor calibration due to the ubiquitous presence of known data in real-world fine-tuning, which appears harmful for calibration. Specifically, data aligned with LLMs' prior knowledge would induce overconfidence, while new knowledge improves calibration. Our findings expose a tension: LLMs’ encyclopedic knowledge, while enabling task versatility, undermines calibration through unavoidable knowledge overlaps. To address this, we propose CogCalib, a cognition-aware framework that applies targeted learning strategies according to the model’s prior knowledge. Experiments across 7 tasks using 3 LLM families prove that CogCalib significantly improves calibration while maintaining performance, achieving an average 57\% reduction in ECE compared to standard fine-tuning in Llama3-8B. These improvements generalize well to out-of-domain tasks, enhancing the objectivity and reliability of domain-specific LLMs, and making them more trustworthy for critical human-AI interaction applications.
\end{abstract}

\section{Introduction}
Large Language Models (LLMs) have enabled powerful domain-specific applications through supervised fine-tuning~\citep{zhuang2023toolqa, imani2023mathprompter, yang2024drhouse}. However, fine-tuning often leads to poor-calibrating LLM, where models' predictive confidence fails to reflect their true performance, manifesting as overconfidence~\citep{achiam2023gpt, zhu2023calibration, shen2024thermometer, yang2023bayesian}. This is particularly concerning in high-stakes scenarios where LLMs' incorrect yet confident predictions could lead to reliability and trustworthiness issues, such as medical diagnosis~\citep{xu2024deepcrbp,zhang2024multi,wei2024dmfvae} or safety-critical domain~\citep{sarabadani2019detection}.
 
\begin{figure}[t]
    \centering
    \includegraphics[width=\linewidth]{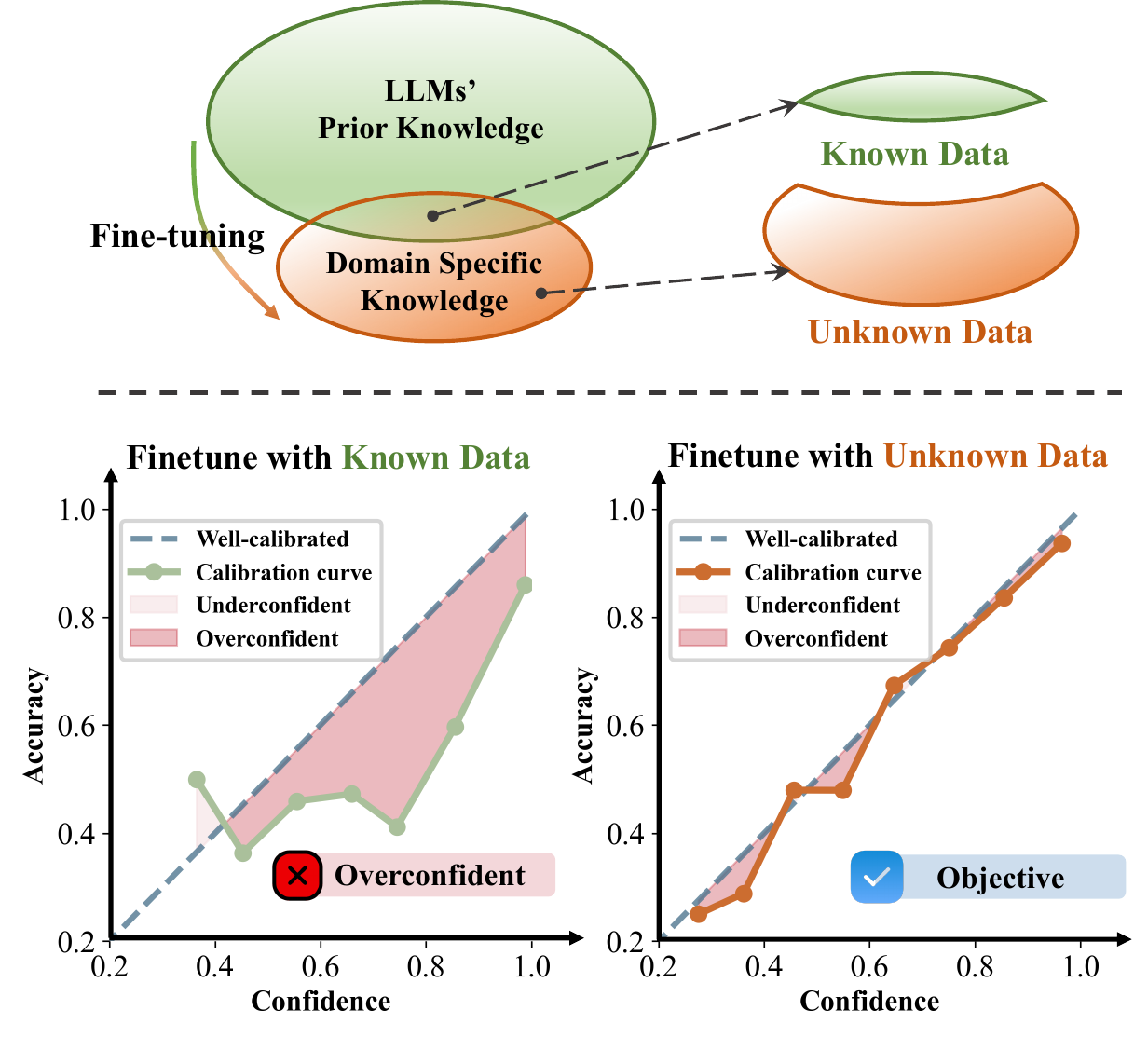}
    \caption{LLMs' prior knowledge leads to poor calibration. As LLMs grow stronger, lots of domain-specific fine-tuning data inevitably overlaps with the LLMs' prior knowledge. We reveal that data aligned with the model's prior knowledge (i.e., \textit{known data}) tend to cause overconfidence, while data exhibiting bias (i.e., \textit{unknown data}) contribute to better alignment between confidence and accuracy, resulting in more objective predictions.}
    \label{fig:overview}
\end{figure}

Prior studies~\citep{mukhoti2020calibrating,wei2022mitigating,pmlr-v70-guo17a} investigating the causes of poor calibration mainly focus on simple models (ResNet) trained from scratch, where prior knowledge is absent. However, in the fine-tuning paradigm of LLMs,  the training data is typically domain-incremental~\citep{shi2024continual}, encompassing both knowledge aligned with the pre-training corpus and novel domain-specific information~\citep{gururangan2020don}. This \textit{knowledge bias} between LLMs' prior knowledge and fine-tuning knowledge has been shown as a critical factor affecting model adaptation~\citep{gekhman2024does,kung2023active,huang2024mitigating,seedat2023curated,chen2024beyond}. Therefore, we try to extend previous research by investigating the underlying mechanisms of poor calibration specifically in the context of fine-tuning, particularly considering the impact of models' prior knowledge.

We reveal that LLMs' extensive prior knowledge, while enabling remarkable few-shot generalization, paradoxically causes their poor calibration in fine-tuning paradigms. Through empirical analysis,
we discover that during fine-tuning, data aligned with the model's prior knowledge (i.e. \textit{known data}) tend to cause overconfidence, while data exhibiting knowledge bias (i.e. \textit{unknown data}) contribute to better calibration as shown in Figure~\ref{fig:overview}. This disparity stems from the distinct learning dynamics between known and unknown data: the model quickly assimilates known data, leading to continued confidence growth even after accuracy plateaus. In contrast, unknown data, inherently more challenging for LLMs to learn~\citep{zhang2024dissecting, gekhman2024does}, results in more synchronized increases in both accuracy and confidence. This phenomenon becomes increasingly problematic as LLMs' prior knowledge expands, making it nearly impossible to avoid overlap between fine-tuning data and their prior knowledge. 

However, existing approaches~\citep{yang2023bayesian, shen2024thermometer, liu2023litcab} are insufficient to handle this issue, primarily because they rely on post-hoc calibration methods. They typically introduce additional learnable modules after fine-tuning to reconstruct the mapping between model outputs and probabilities, which incurs extra computational overhead during deployment. On the other hand, the influence of LLMs' prior knowledge on calibration provides a promising opportunity to address poor calibration during fine-tuning.

Therefore, we introduce CogCalib, a real-time fine-tuning calibration framework compatible with various training-based calibration methods. Specifically, CogCalib dynamically evaluates knowledge bias during fine-tuning and applies targeted learning strategies accordingly, regulating confidence fitting and maintaining task learning. Moreover, CogCalib introduces no additional computational overhead during deployment.

We conduct comprehensive experiments across 7 commonly used downstream tasks (including multiple-choice and open-ended QA tasks) using 3 popular LLM families, to demonstrate CogCalib's effectiveness. CogCalib successfully preserves fine-tuning performance while achieving substantial improvements in calibration across all tasks and models, without incurring additional computational overhead during deployment. For instance, Llama3-8B achieves average ECE reductions of 55.92\% and 65.02\% compared to TS and SFT on multiple-choice QA tasks. Notably, these improvements generalize well to out-of-domain tasks, indicating that models trained with CogCalib consistently demonstrate enhanced objectivity across diverse scenarios. The main contributions of our work can be summarized as follows:

\begin{itemize}
\item As far as we know, we are the first to reveal the neglected negative impacts of LLMs' prior knowledge on calibration during fine-tuning. Specifically, data aligned with the model's prior knowledge tends to induce overconfidence, while new knowledge is beneficial for calibration.
\item We propose CogCalib, a real-time calibration framework that employs distinct learning strategies for data with different knowledge biases during fine-tuning, aiming to achieve more objective fine-tuning.
\item We conduct extensive experiments on domain-specific multiple-choice and open-ended QA tasks with multiple models, using different fine-tuning methods, which demonstrate the effectiveness and generality of CogCalib in enhancing calibration. 
\end{itemize}

\section{Related Works}
\subsection{Confidence Calibration}
Confidence calibration methods can be categorized into three main approaches~\citep{gawlikowski2023survey}: post-processing adjustments~\citep{guo2017calibration}, training-based optimization~\citep{szegedy2016rethinking}, and uncertainty estimation~\citep{lakshminarayanan2017simple}. For LLMs specifically, recent efficient post-processing techniques have emerged, including Bayesian LoRA~\citep{yang2023bayesian}, LLM-oriented temperature scaling~\citep{shen2024thermometer}, and distribution adjustment methods~\citep{liu2023litcab}. While these approaches address calibration computational complexity, they do not investigate the underlying causes of calibration degradation. Previous studies have identified negative log-likelihood (NLL) overfitting as a key factor in poor calibration~\citep{mukhoti2020calibrating,wei2022mitigating,pmlr-v70-guo17a}. However, these findings were based on models without prior knowledge, whereas the unique pre-trained nature of LLMs~\citep{shi2024continual} necessitates a fresh examination of calibration.
\subsection{Impact of Knowledge Bias in Fine-tuning}
Fine-tuning LLMs presents several critical challenges, including hallucinations~\citep{10.1145/3703155}, generalization problems~\citep{he-etal-2021-analyzing}, and calibration degradation~\citep{zhu2023calibration}, with the bias between LLMs' prior knowledge and fine-tuning knowledge emerging as a contributing factor~\citep{kung2023active,huang2024mitigating,seedat2023curated,yang-etal-2024-self}. Gekhman et al.~\citep{gekhman2024does} demonstrate that introducing new knowledge during fine-tuning can trigger hallucinations. Effective generalization can be achieved through knowledge selection strategies based on knowledge bias~\citep{albalak2024survey,chen2024beyond}. Regarding calibration, the impact of knowledge bias during fine-tuning warrants further investigation.

\section{Prior Knowledge Affects Calibration?}
\label{sec:3}
\begin{figure}
    \centering
    \includegraphics[width=\linewidth]{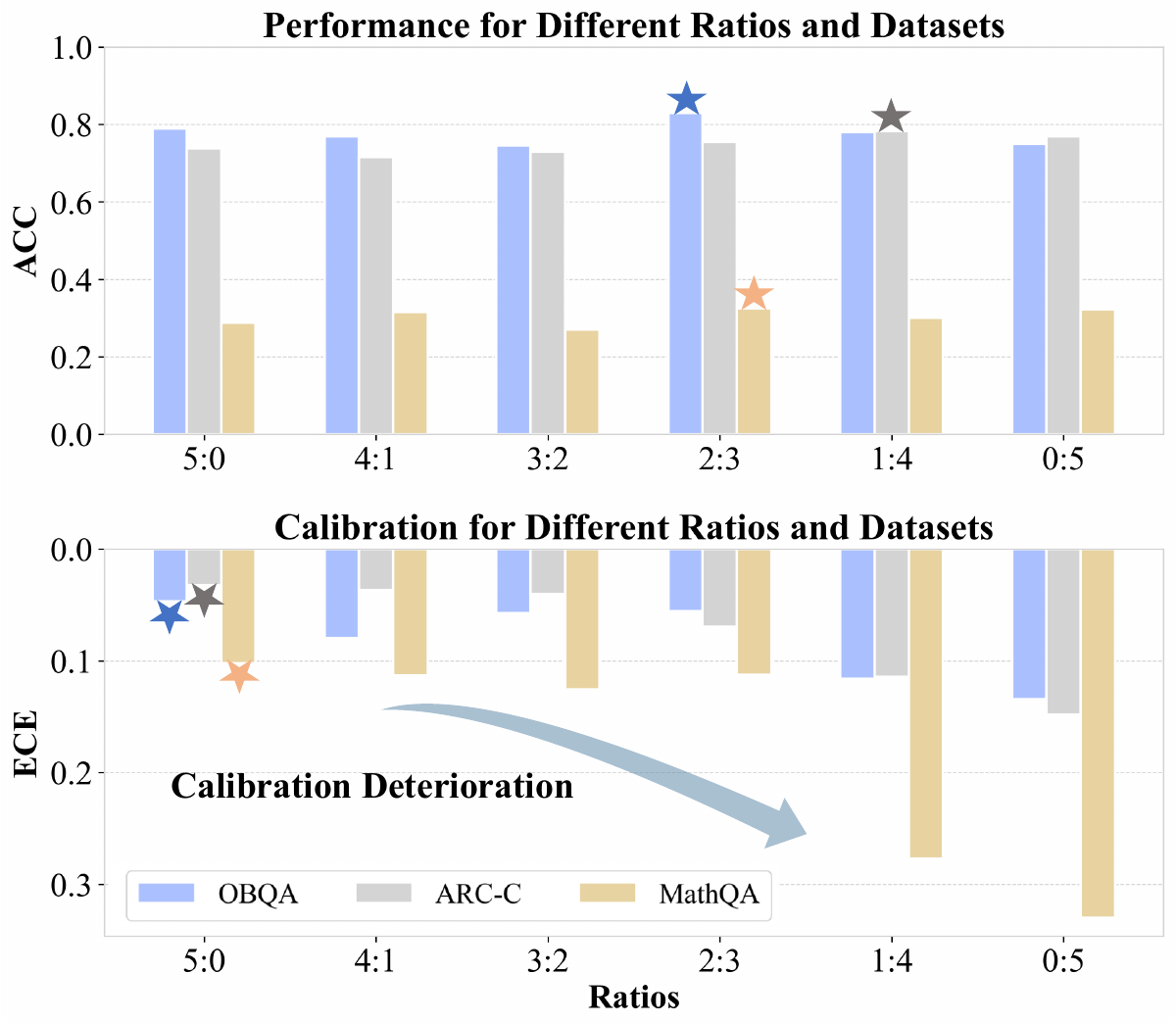}
    \caption{Accuracy and ECE of Llama3-8B fine-tuned with different knowledge biases. We fine-tune Llama3-8B using OBQA, with ARC-C and MathQA as OOD tests. The ratio varies from 5:0 to 0:5 (unknown data:known data), with equal dataset sizes. Calibration deteriorates as the knowledge bias lowers, while a higher knowledge bias helps improve calibration.}
    \label{fig:distribution_impact}
\end{figure}
In this section, we investigate how LLMs' prior knowledge affects calibration during fine-tuning. To quantify the overlap between fine-tuning data and the model's prior knowledge, we first reintroduce the concept of \textit{knowledge bias}, which represents the discrepancy between the model's prior knowledge domain and the downstream task knowledge domain. Following the framework proposed by SliCK~\citep{gekhman2024does} (details shown in Appendix~\ref{appendix:SliCK}), we categorize the data into two distinct types: \textit{known data} that aligns with the model's prior knowledge, and \textit{unknown data} that deviates from this knowledge base. Finally, we simulate varying knowledge bias by adjusting the ratio between unknown and known data in the fine-tuning dataset, where a higher proportion of known data indicates a lower knowledge bias (i.e., greater alignment with prior knowledge). The details of data construction are shown in Appendix~\ref{appendix:data_construct}.

\subsection{Minimal Bias, Maximal Overconfidence}
\label{sec:3.1}
To simulate varying knowledge biases, we construct fine-tuning datasets with six ratios of unknown to known data in OBQA. While Figure~\ref{fig:distribution_impact} reveals irregular performance trends across knowledge bias levels, the calibration exhibits a clear directional pattern: \textbf{lower knowledge bias consistently degrades calibration, whereas higher bias improves it}, a phenomenon persistent across both in-domain and out-of-domain. Notably, the introduction of even a small fraction of known data leads to calibration deterioration (from pure unknown data to 4:1 ratio). This suggests that the model's pre-existing knowledge dominance for calibration begins immediately upon exposure to aligned data. We observed the same phenomenon across other models and datasets (further experimental results and analyses are provided in the Appendix~\ref{appendix:more_of_sec3}).

\begin{figure}
    \centering
    \includegraphics[width=\linewidth]{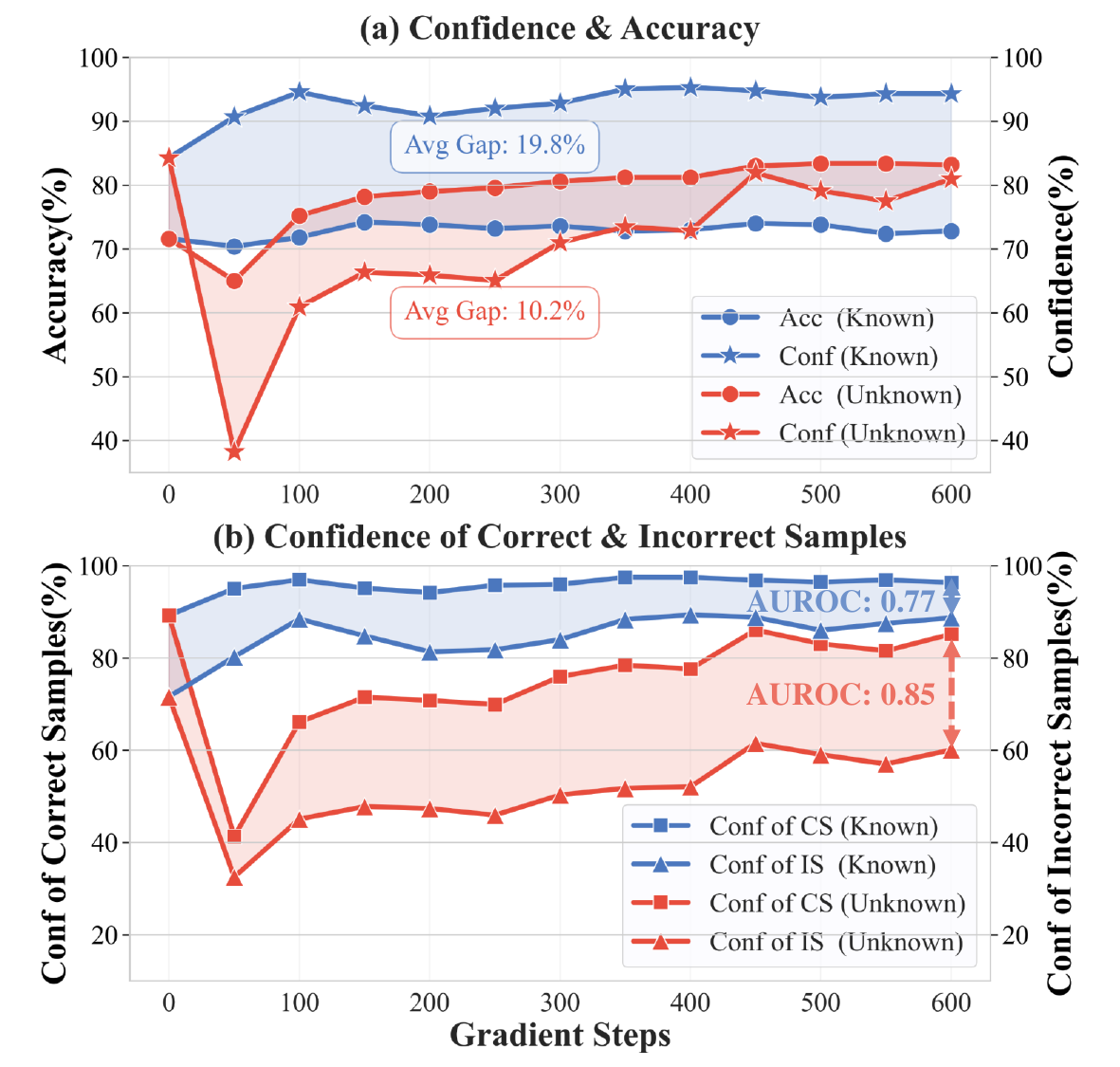}
    \caption{(a) Accuracy and confidence of Llama3-8B during fine-tuning on known and unknown data sampled from OBQA. The asymmetric fitting rates between accuracy and confidence in known fine-tuning result in model overconfidence. Conversely, unknown fine-tuning exhibits synchronized fitting of both, minimizing their disparity. (b) Average confidence of correct and incorrect predictions. Unknown fine-tuning yields distinct confidence separation between correct and incorrect samples, facilitating OOD detection.}
    \label{fig:verify-id-conf}
\end{figure}

Furthermore, our tracking for accuracy and confidence reveals divergent learning dynamics: in low-bias fine-tuning (Figure~\ref{fig:verify-id-conf}a), accuracy of OBQA test set plateaus early (200 steps) while confidence escalates continuously, creating widening calibration error. Conversely, high-bias conditions maintain synchronized accuracy-confidence growth, minimizing discrepancies — a pattern potentially rooted in gradual new knowledge assimilation~\citep{zhang2024dissecting, gekhman2024does}. 
The different confidence patterns persist in OOD detection~\citep{10.1007/978-3-030-87735-4_12} (Figure~\ref{fig:verify-id-conf}b): low-bias models show compressed confidence distributions for correct and incorrect predictions (both correct/incorrect > 85\%), whereas high-bias models develop discriminative confidence gaps (AUROC 0.85 vs 0.77 at step 600), enhancing OOD detection. More results, including standard deviations are shown in Appendix \ref{appendix:more_of_sec3}.

These findings collectively demonstrate how LLMs' prior knowledge induces poor calibration: \textbf{pre-existing knowledge enables rapid confidence inflation on aligned data, while insufficient exposure to new knowledge prevents calibration improvement — a harmful interaction amplified by the near-ubiquitous presence of known data in the real-world fine-tuning.} Additionally, we examine this phenomenon in realistic scenarios (details are shown in Figure~\ref{fig:verify-ece} of Appendix~\ref{appendix:more_of_sec3}).

\subsection{Analysis and a Potential Solution}
\label{sec:3.2}
An intuitive explanation is that known samples closely align with pre-trained models' prior distribution, while unknown samples represent the target distribution. Therefore, fine-tuning in low-bias scenarios leads to rapid confidence overfitting. In contrast, high-bias fine-tuning requires the model to adjust its decision boundaries to accommodate new distributions, resulting in better calibration. 

A potential solution is to increase the bias between the fine-tuning data and the model's prior knowledge, specifically by eliminating known data. However, we reveal that simple bias adjustment through data removal is insufficient, as it fails to consistently improve calibration performance across different datasets.
\begin{figure}[!htbp]
    \centering
    \includegraphics[width=\linewidth]{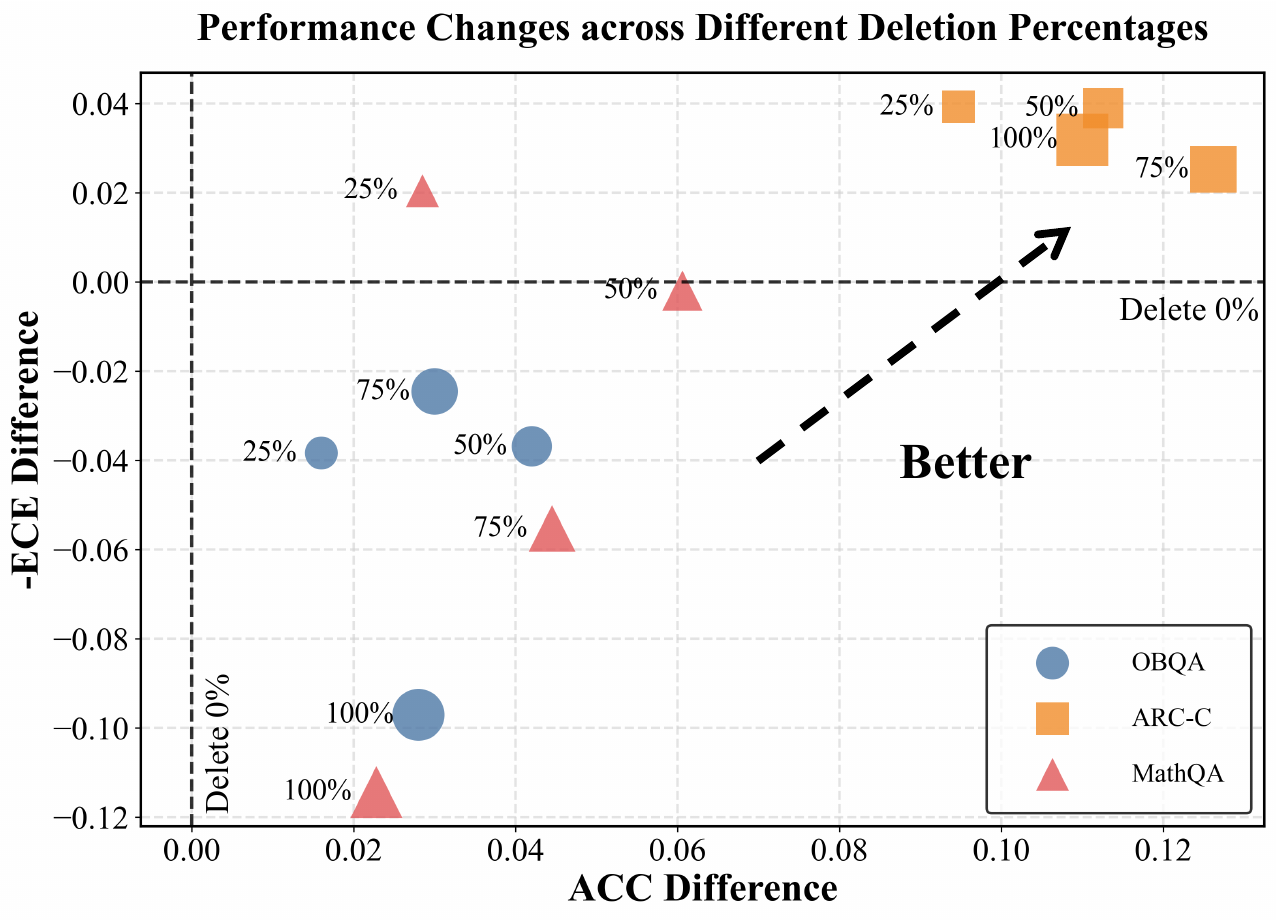}
    \caption{Differences in ACC and ECE compared to baseline (delete 0\%) under various percentages of known data deletion. The results from in-domain tests indicate that simple bias adjustment fails to achieve consistent calibration improvements across all tasks.}
    \label{fig:delete}
\end{figure}
As shown in Figure~\ref{fig:delete}, removing 25\% of low-bias data improves calibration in ARC-C but degrades it in OBQA. This inconsistency may stem from the inherent characteristics of different datasets, making it challenging to find a universal optimal adjustment ratio. Additionally, accuracy consistently improves with known data reduction, aligning with findings new knowledge enhances task performance~\citep{kung2023active,swayamdipta-etal-2020-dataset}. This necessitates methods that decouple knowledge retention from calibration, preserving fine-tuning performance while improving calibration.

\section{Cognition-aware Calibration}
\label{sec:4}
In this section, we propose CogCalib, a \textbf{Cog}nition-aware \textbf{Calib}ration framework for fine-tuning, designed to achieve an optimal balance between fine-tuning performance and calibration. CogCalib is motivated by the above observation that known and unknown data exhibit distinct fitting characteristics during the fine-tuning process, necessitating different learning strategies. To develop an effective solution, we mainly address two challenges: 
(1) \textit{How to evaluate knowledge bias, particularly as the model's internal states are continuously evolving?} 
(2) \textit{What specific learning strategies should be applied to achieve objective fine-tuning?}

\begin{figure}[h]
    \centering
    \includegraphics[width=\linewidth]{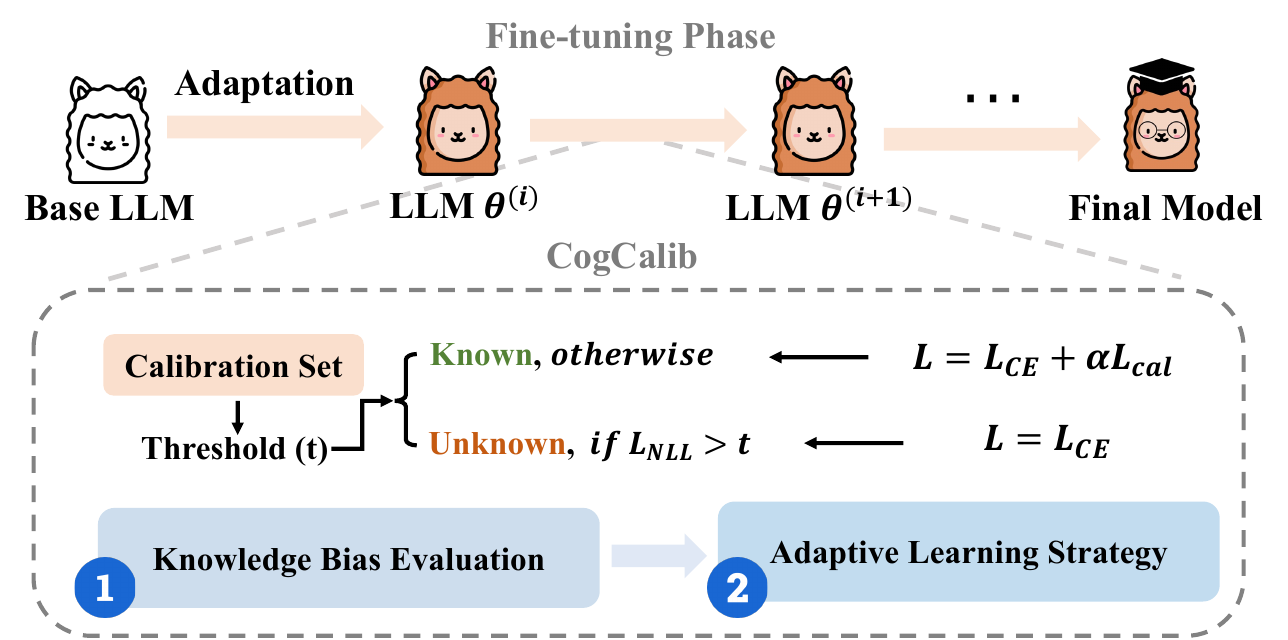}
    \caption{CogCalib's framework. CogCalib dynamically assesses knowledge bias during training through NLL, employing customized learning strategies with distinct loss functions to enhance calibration. Additionally, CogCalib incorporates a style adaptation process to improve the knowledge bias evaluation performance.}
    \label{fig:cogcalib}
\end{figure}

\subsection{Knowledge Bias Evaluation}
In section~\ref{sec:3}, we evaluate knowledge bias based on the correctness of the model's output through multiple inferences. However, during training, this method faces limitations due to the inability to perform multiple sampling iterations. 

Therefore, we propose a more efficient method for knowledge bias assessment based on negative log-likelihood (NLL). Our hypothesis posits that known data aligns with the pre-trained model's prior knowledge bias, while unknown data represents novel information from the target distribution. From a distributional perspective, as shown in Equation (\ref{eq:nll}), known data should exhibit lower NLL values compared to unknown data:
\begin{equation}
\label{eq:nll}
\mathcal{L}_{\mathrm{NLL}}=\mathbb{E}_{\mathbf{q}}\left[\mathbf{p}\right]=-\sum_{k=1}^Cq_k\log p_k,  
\end{equation}
where $\mathbf{q}$ is the target one-hot distribution, $\mathbf{p}$ is the vocabulary distribution output by LLM, and $p_k$ and $q_k$ represent the probabilities for the k-th class in $\mathbf{p}$ and $\mathbf{q}$, respectively.  Based on this, we evaluate knowledge bias during training, using Equation (\ref{judge_k_unk}),
\begin{equation}
\label{judge_k_unk}
\mathbf{I}(\mathbf{p}, \mathbf{q}) =
\begin{cases}
1, & \text{if } \mathcal{L}_{\mathrm{NLL}} \leq t \\
0, & \text{otherwise}
\end{cases},
\end{equation}
where $t$ denotes the threshold, and $I=1$ indicates the model has already mastered this knowledge. In addition, $t$ requires adjustment to accommodate the evolving knowledge distribution during training. To solve this, we establish a calibration set to identify the optimal $t$ according to Algorithm~\ref{alg:threshold} (details of the calibration set are shown in Appendix~\ref{Appendix:dataset}).

\begin{algorithm}[t]
\caption{Adaptive Threshold Update}\label{alg:threshold}
\KwData{Calibration set $S_{c}=\{(x_i,y_i)\}_{i=1}^{N}$, grid size $M$}
\KwResult{Updated threshold $t^{\ast}$}
\ForEach{update step (\emph{e.g.}, every epoch)}{
    \For{$i\leftarrow1$ \KwTo $N$}{
        $(c_i,n_i)\leftarrow\text{Inference}(x_i)$\; 
        \tcp{$c_i{=}1$ if correct else $0$}
        \tcp{$n_i=-\log p_{\theta}(y_i|x_i)$}
    }
    \tcp{Generate candidate thresholds}
    $\mathcal{T}\leftarrow\text{linspace}\bigl(\min n_i,\;\max n_i,\;M\bigr)$\;
    \tcp{Grid-search for optimal $t$}
    $t^{\ast}\leftarrow\arg\max_{t\in\mathcal{T}}\bigl(\text{TPR}(t)+\text{TNR}(t)\bigr)$\;
    \tcp{$\text{TPR}$: true positive rates}
    \tcp{$\text{TNR}$: true negative rates}
}
\end{algorithm}

However, the discrepancy in LLMs' linguistic style and label formats prevents NLL from accurately assessing knowledge bias. Thus, the model requires a style adaptation process for calculating the initial threshold $t_0$, based on findings~\citep{zhang2024dissecting, mai2024fine} that LLMs rapidly adapt to downstream task syntax during early fine-tuning (details are shown in Appendix~\ref{appendix:adaptation}). 

\subsection{Adaptive Learning Strategy}
Our previous analysis reveals that during fine-tuning, model confidence increases rapidly for known data, while unknown data contributes positively to both calibration and downstream task performance. Therefore, we moderate confidence fitting for known data while preserving the learning dynamics for unknown data as Equation (\ref{eq:loss}),
\begin{equation}
\label{eq:loss}
    \mathcal{L}=\mathcal{L}_{\mathrm{CE}}+\mathbf{I}(\mathbf{p},\mathbf{q})\cdot \alpha \mathcal{L}_{cal},
\end{equation}
where $\alpha$ represents the regularization strength and $ \mathcal{L}_{cal}$ denotes the calibration loss during training. The calibration term could be Label Smoothing (LS)~\citep{szegedy2016rethinking}, Margin-based Label Smoothing (MbLS)~\citep{liu2022devil}, or ECP~\citep{pereyra2017regularizing}, which have been proved to be helpful for confidence overfitting (details shown in Appendix~\ref{Appendix:cal_methods}). In CogCalib, we call these methods CoLS, CoMbLS, and CoECP.

\subsection{Integrated Framework} 
Building on our previous analysis, we propose an integrated framework aiming to achieve objective fine-tuning. Figure~\ref{fig:cogcalib} illustrates the architecture of CogCalib. Initially, LLM undergoes style adaptation to align with the grammatical patterns of downstream tasks. Subsequently, CogCalib dynamically assesses knowledge bias using NLL and adaptive $t$. For low-bias data, we incorporate a calibration term to mitigate confidence overfitting, while cross-entropy loss is applied to new knowledge to maintain task alignment.

\begin{table}[h]
\centering
\small
\setlength{\tabcolsep}{4.2pt} 
\begin{tabular}{ccccc}
\toprule 
\textbf{Type} & \textbf{Dataset} & \textbf{Accuracy} & \textbf{TPR} & \textbf{TNR} \\
\midrule
\multirow{5}{*}{\textbf{Multi-Choice}} 
& \textbf{OBQA} & 99.44 & 99.44 & 99.52 \\
& \textbf{ARC-C} & 99.51 & 99.54 & 99.15 \\
& \textbf{WG-S} & 98.83 & 98.77 & 99.52 \\
& \textbf{WG-M} & 99.13 & 99.10 & 99.55 \\
& \textbf{BoolQ} & 98.69 & 98.63 & 98.21 \\
\midrule
\multirow{2}{*}{\textbf{Open-End}}
& \textbf{HotpotQA} & 83.64 & 79.43 & 90.59 \\
& \textbf{MedMCQA}  & 83.69 & 79.77 & 87.78 \\
\bottomrule
\end{tabular}
\caption{Accuracy, True Positive Rate (TPR), and True Negative Rate (TNR) for identifying known/unknown data using NLL in the fine-tuning process of Llama3-8B.}
\label{tab:nll_classify_acc}
\end{table}
\section{Experiments}

In this section, we will evaluate the universality of CogCalib across 3 aspects: diverse datasets (from multiple-choice to open-ended), various LLM families and sizes, and different fine-tuning approaches (LoRA and FFT). Complete experimental results are presented in Appendix~\ref{appenix:exp}, and hyperparameters settings are shown in Appendix~\ref{appendix:hyper}.

\begin{table*}[t]
\centering
\small
\setlength{\tabcolsep}{4pt} 
\begin{tabular}{cccccc|ccc} 
		\toprule 
		\textbf{Dataset} & \textbf{Metric} & \textbf{Vanilla SFT} & \textbf{MCD} & \textbf{Ensemble} & \textbf{TS} & \textbf{CoLS} ($\Delta$TS) & \textbf{CoMbLS} ($\Delta$TS) & \textbf{CoECP} ($\Delta$TS) \\
		\midrule
		\multirow{2}{*}{\textbf{OBQA}} 
        & \textbf{ACC$\uparrow$} & 84.80 & 83.60 & \textbf{88.00} & 84.80 & 85.60 (\textcolor{deepgreen}{+0.80}) & 86.20 (\textcolor{deepgreen}{+1.40}) & 86.20 (\textcolor{deepgreen}{+1.40}) \\
		& \textbf{ECE$\downarrow$} & 11.20 & 9.10 & 6.02 & 9.90 & \textbf{2.50} (\textcolor{deepgreen}{-7.40}) & 3.70 (\textcolor{deepgreen}{-6.20}) & 7.30 (\textcolor{deepgreen}{-2.60}) \\
		
		\midrule
		\multirow{2}{*}{\textbf{ARC-C}} 
        & \textbf{ACC$\uparrow$} & 81.40 & 80.70 & 81.14 & 81.40 & 81.60 (\textcolor{deepgreen}{+0.20}) & \textbf{81.70} (\textcolor{deepgreen}{+0.30}) & 81.60 (\textcolor{deepgreen}{+0.20}) \\
		& \textbf{ECE$\downarrow$} & 16.50 & 13.80 & 13.08 & 12.30 & 4.80 (\textcolor{deepgreen}{-7.50}) & \textbf{4.20} (\textcolor{deepgreen}{-8.10}) & 7.40 (\textcolor{deepgreen}{-4.90}) \\
		
		\midrule
		\multirow{2}{*}{\textbf{WG-S}} 
        & \textbf{ACC$\uparrow$} & 78.00 & 78.21 & 80.27 & 78.00 & 80.10 (\textcolor{deepgreen}{+2.10}) & 79.20 (\textcolor{deepgreen}{+1.20}) & \textbf{80.30} (\textcolor{deepgreen}{+2.30}) \\
		& \textbf{ECE$\downarrow$} & 20.50 & 17.77 & 14.81 & 15.40 & 8.90 (\textcolor{deepgreen}{-6.50}) & 9.40 (\textcolor{deepgreen}{-6.00}) & \textbf{7.00} (\textcolor{deepgreen}{-8.40}) \\
		
		\midrule
		\multirow{2}{*}{\textbf{WG-M}} 
        & \textbf{ACC$\uparrow$} & 84.50 & 84.37 & \textbf{85.16} & 84.50 & 84.50 (\textcolor{deepgreen}{+0.0}) & 84.70 (\textcolor{deepgreen}{+0.20}) & 84.70 (\textcolor{deepgreen}{+0.20}) \\
		& \textbf{ECE$\downarrow$} & 14.80 & 13.51 & 10.09 & 11.50 & 4.10 (\textcolor{deepgreen}{-7.40}) & 3.10 (\textcolor{deepgreen}{-8.40}) & \textbf{1.00} (\textcolor{deepgreen}{-10.50}) \\
		
		\midrule
		\multirow{2}{*}{\textbf{BoolQ}} 
        & \textbf{ACC$\uparrow$} & 90.09 & 90.15 & \textbf{90.86} & 90.09 & 90.15 (\textcolor{deepgreen}{+0.06}) & 89.63 (\textcolor{deepred}{-0.46}) & 89.54 (\textcolor{deepred}{-0.55}) \\
		& \textbf{ECE$\downarrow$} & 9.54 & 8.95 & 6.69 & 7.70 & \textbf{1.97} (\textcolor{deepgreen}{-5.73}) & 2.36 (\textcolor{deepgreen}{-5.34}) & 7.68 (\textcolor{deepgreen}{-0.02}) \\
		
		\bottomrule
\end{tabular}
\caption{Comparison of our method's performance against baselines on in-domain (ID) datasets. Results are evaluated on Llama3-8B model fine-tuned by LoRA on 5 widely used domain-specific datasets. We integrate LS, MbLS, and ECP as calibration terms in CogCalib, resulting in 3 variants: CoLS, CoMbLS, and CoECP.}
\label{tab:id}
\end{table*}

\subsection{Experimental Setup}
\noindent \textbf{Datasets.}
To ensure the universality of CogCalib, we select a wide range of tasks, including HotpotQA~\citep{yang2018hotpotqa} MedMCQA~\citep{pmlr-v174-pal22a} for open-ended QA tasks, while utilizing OpenBookQA (OBQA)~\citep{OpenBookQA2018}, ARC-Challenge (ARC-C)~\citep{clark2018think}, Winogrande-small (WG-S), Winogrande-medium (WG-M)~\citep{sakaguchi2021winogrande} and BoolQ~\citep{clark2019boolq} for multiple-choice QA scenarios. Additionally, we extend our evaluation of CogCalib to various OOD tasks, including MMLU~\citep{hendryckstest2021} and ARC-E~\citep{clark2018think}. See Appendix~\ref{Appendix:dataset} for more details.

\noindent \textbf{Models.}
We validate CogCalib across models of diverse families and scales, including Llama3-8B, Llama2-13B, Mistral-7B, and Qwen2.5-7B. All models employed in our experiments are instruction-tuned variants of their respective base models.

\noindent \textbf{Evaluation Metrics.}
In addition to evaluating the accuracy of fine-tuned models, we also select ECE with a bin size of 10 to assess calibration. See Appendix~\ref{appendix:ece} for more details of ECE.

\noindent \textbf{Baselines.}
We consider 4 baseline methods: (1) \textit{Vanilla SFT}: We use standard LoRA or FFT as a lower performance bound. (2) \textit{MC-Dropout (MCD)}~\citep{gal2016dropout}: We use a dropout rate of 0.02 during fine-tuning and perform sampling 4 times. (3) \textit{Deep Ensemble (Ensemble)}~\citep{Lakshminarayanan_Pritzel_Blundell_2016}: We use 3 fine-tuned LLMs. (4) \textit{Temperature Scaling (TS)}~\citep{guo2017calibration}: The optimal temperature is calculated on the ID validation set and applied to both the ID and OOD datasets. See Appendix~\ref{appendix:ts} for more details.

\subsection{Main Results}
In this section, we validate the effectiveness of CogCalib through comprehensive experiments. First, we demonstrate the validity of using NLL for assessing knowledge bias, which enhances the interpretability of our framework. Subsequently, we evaluate CogCalib's performance on both multiple-choice and open-ended tasks, showing that it not only maintains fine-tuning performance but also significantly improves calibration.

\subsubsection{Effectiveness Evaluation of Knowledge Bias via NLL}
In this section, we aim to validate the effectiveness of using NLL for evaluating knowledge bias. Table~\ref{tab:nll_classify_acc} presents accuracy in distinguishing between unknown/known data using NLL during training (average accuracy throughout the training process). The high accuracy proves NLL-based method aligns well with SliCK~\citep{gekhman2024does}, validating NLL as an effective knowledge bias evaluation metric. Moreover, we demonstrate our threshold calculation method outperforms alternative approaches in Appendix~\ref{appendix:sense_to_threshold}.

\subsubsection{Calibration of Multi-Choice Task}
To verify CogCalib's robustness and generalizability, our evaluation for CogCalib consists of two dimensions: in-domain performance assessment and out-of-domain evaluation.

\begin{table*}[t]
\centering
\small
\setlength{\tabcolsep}{2.5pt} 
\begin{tabular}{cc|c|cc|cccc}
		\toprule 
		\multirow{2}{*}{\textbf{Metric}} &\multirow{2}{*}{\textbf{Methods}} &\multicolumn{1}{c}{\textbf{In Domain}} &  \multicolumn{2}{|c}{\textbf{Smaller Distribution Shift}}&
		\multicolumn{4}{|c}{\textbf{Larger Distribution Shift}}
		\\& & OBQA &ARC-C  & ARC-E & Business & Culture & History & Psychology \\
		\midrule
\multirow{8}{*}{\textbf{ECE}$\downarrow$}&
\textbf{Vanilla SFT}& 11.20 & 18.00 & 13.50 & 18.40 & 17.61 & 19.22 & 23.38 \\
& \textbf{MCD} & 9.10 & 14.56 & 11.11 & 13.54 & 15.70 & 17.87 & 20.42 \\
& \textbf{Ensemble} & 6.02 & 14.59 & 8.92 & 14.09 & 15.33 & 15.99 & 18.76 \\
& \textbf{TS} & 9.90 & 15.90 & 10.40 & 16.10 & 16.70 & 17.40 & 21.30 \\
\cmidrule{2-9} 
& \textbf{CoLS} ($\Delta$TS) & \textbf{2.50} (\textcolor{deepgreen}{-7.4}) & 7.50 (\textcolor{deepgreen}{-8.4}) & 2.40 (\textcolor{deepgreen}{-8.0}) & 9.80 (\textcolor{deepgreen}{-6.3}) & 10.30 (\textcolor{deepgreen}{-6.4}) & 12.07 (\textcolor{deepgreen}{-5.3}) & 14.75 (\textcolor{deepgreen}{-6.6}) \\	
& \textbf{CoMbLS} ($\Delta$TS) & 3.70 (\textcolor{deepgreen}{-6.2}) & 5.80 (\textcolor{deepgreen}{-10.1}) & \textbf{1.40} (\textcolor{deepgreen}{-9.0}) & 8.20 (\textcolor{deepgreen}{-7.9}) & 9.48 (\textcolor{deepgreen}{-7.2}) & 9.83 (\textcolor{deepgreen}{-7.6}) & 14.41 (\textcolor{deepgreen}{-6.9}) \\	
& \textbf{CoECP} ($\Delta$TS) & 7.30 (\textcolor{deepgreen}{-2.6}) & \textbf{2.80} (\textcolor{deepgreen}{-13.1}) & 4.90 (\textcolor{deepgreen}{-5.5}) & \textbf{3.80} (\textcolor{deepgreen}{-12.3}) & \textbf{3.46} (\textcolor{deepgreen}{-13.2}) & \textbf{6.27} (\textcolor{deepgreen}{-11.1}) & \textbf{9.51} (\textcolor{deepgreen}{-11.8}) \\
\midrule 

\multirow{8}{*}{\textbf{ACC}$\uparrow$}&
\textbf{Vanilla SFT}& 84.80 & 79.10 & 84.10 & 79.20 & 79.52 & 77.63 & 73.47 \\
& \textbf{MCD} & 83.60 & 78.92 & 84.22 & 80.78 & 79.22 & 76.24 & 73.38 \\
& \textbf{Ensemble} & \textbf{88.00} & 79.35 & \textbf{87.37} & 80.32 & 79.52 & 78.39 & \textbf{75.11} \\
& \textbf{TS} & 84.80 & 79.10 & 84.10 & 79.20 & 79.52 & 77.63 & 73.47 \\
\cmidrule{2-9} 
& \textbf{CoLS} ($\Delta$TS)& 85.60 (\textcolor{deepgreen}{+1.8}) & 79.30 (\textcolor{deepgreen}{+0.2}) & 86.30 (\textcolor{deepgreen}{+2.2}) & 79.40 (\textcolor{deepgreen}{+0.2}) & 78.92 (\textcolor{deepred}{-0.6}) & 76.24 (\textcolor{deepred}{-1.4}) & 73.64 (\textcolor{deepgreen}{+0.2}) \\	
& \textbf{CoMbLS} ($\Delta$TS) & 86.20 (\textcolor{deepgreen}{+2.4}) & \textbf{80.00} (\textcolor{deepgreen}{+0.9}) & 86.70 (\textcolor{deepgreen}{+2.6}) & \textbf{81.70} (\textcolor{deepgreen}{+2.5}) & 80.12 (\textcolor{deepgreen}{+0.6}) & \textbf{78.92} (\textcolor{deepgreen}{+1.3}) & 74.50 (\textcolor{deepgreen}{+1.0}) \\	
& \textbf{CoECP} ($\Delta$TS) & 86.20 (\textcolor{deepgreen}{+2.4}) & 79.00 (\textcolor{deepred}{-0.1}) & 84.60 (\textcolor{deepgreen}{+0.5}) & 80.80 (\textcolor{deepgreen}{+1.6}) & \textbf{81.02} (\textcolor{deepgreen}{+1.5}) & 77.96 (\textcolor{deepgreen}{+0.3}) & 74.50 (\textcolor{deepgreen}{+1.0})\\
\bottomrule
\end{tabular}
\caption{Comparison of our method's performance against baselines on distribution shift datasets is presented. Results are evaluated on Llama3-8B model which is fine-tuned on the OBQA dataset.}
\label{tab:idood}
\end{table*}

\begin{figure*}[h]
    \centering
    \includegraphics[width=\linewidth]{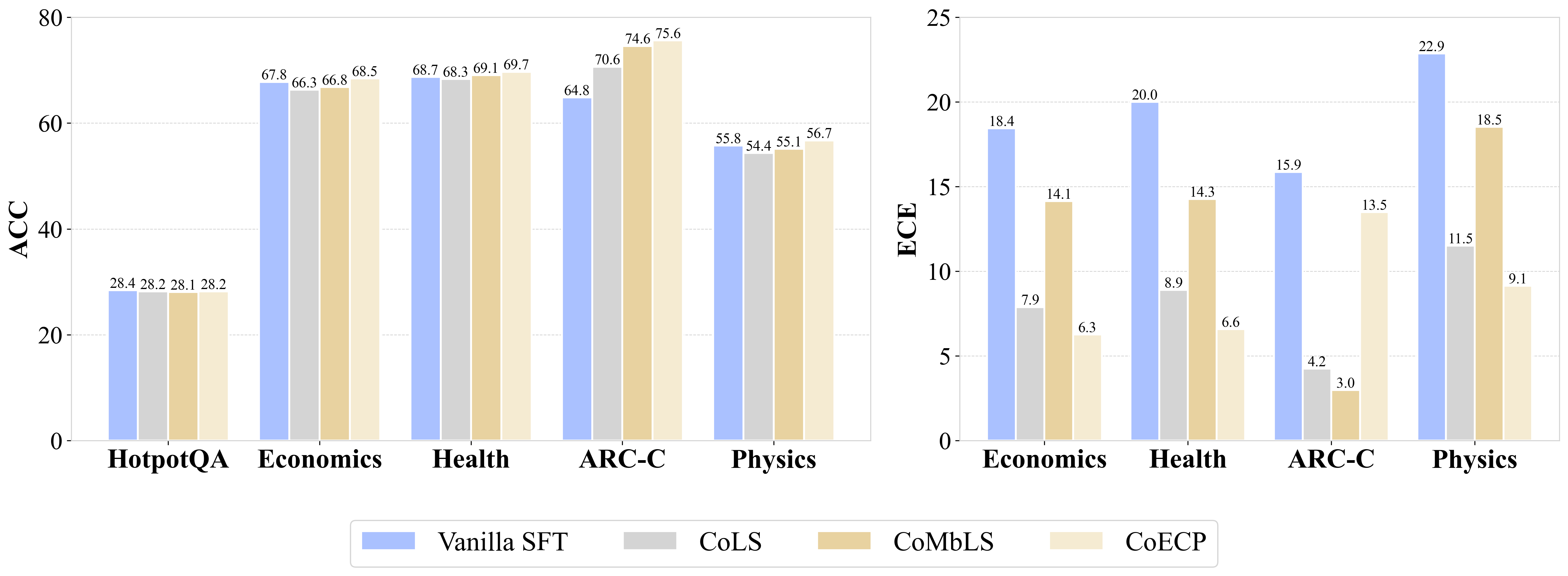}
    \caption{Comparison of our method's performance against baseline approaches on OOD datasets is presented. The results are evaluated on the Llama3-8B model, which is fine-tuned on the open-ended HotpotQA dataset.}
    \label{fig:llama3-hot-ood}
\end{figure*}

\noindent \textbf{In-Distribution Performance.} We first conduct in-domain tests on CogCalib across 5 commonsense reasoning datasets. On the one hand, our cognitive methods maintain competitive accuracy compared to baselines as shown in Table~\ref{tab:id}, such as CoECP achieving 86.20\% and 80.30\% accuracy on OBQA and WG-S datasets, respectively. On the other hand, our cognitive methods achieve substantial calibration improvements across all datasets. These indicate that CogCalib achieves more objective fine-tuning on ID tasks. 

\noindent \textbf{Performance Under Distribution Shift.} 
Real-world applications demand robust model performance across different scenarios. We evaluate CogCalib under various distribution shifts, including ARC-C and ARC-E datasets for smaller shifts, and 4 MMLU subjects (Business, Culture, History, Psychology) for larger domain shifts. Table~\ref{tab:idood} indicates that CogCalib maintains competitive accuracy compared to the baselines under distribution shifts and achieves overall superior ECE. These findings demonstrate the robustness of the CogCalib in tasks with distribution shifts. 

\noindent \textbf{More experiments based on other LLMs.} To validate the generalizability of CogCalib, we also conduct experiments on Mistral-7B, Qwen2.5-7B, and Llama2-13B in Appendix~\ref{appenix:exp}. The results demonstrate that CogCalib consistently achieves significant calibration improvements across these models while maintaining fine-tuning performance, proving the cross-model generalizability of CogCalib.

\noindent \textbf{More experiments based on FFT.} To explore the applicability of CogCalib to other fine-tuning methods, we validate CogCalib with FFT on Llama3-8B, demonstrating its effectiveness beyond LoRA-based approaches (see Appendix~\ref{appendix:fft_ret}).

\subsubsection{Calibration of Open-End Task}
In addition, we evaluate CogCalib on open-ended datasets HotpotQA (experiments on MedMCQA are presented in Appendix~\ref{appendix:llama3-8b}). As shown in Figure~\ref{fig:llama3-hot-ood}, CogCalib maintains accuracy on ID and OOD tasks while providing larger accuracy gains on some datasets, e.g., the CoECP shows a 10.8\% ACC gain over Vanilla SFT on ARC-C. Regarding calibration, our cognitive methods exhibit comprehensive improvement. These further demonstrate the task-agnostic nature of CogCalib.

\subsection{Ablation Study}
\label{subsec:abala}
\noindent \textbf{Comparision to Vanilla and Random Calibration.} In this section, we validate the necessity of employing different learning strategies for known and unknown data within CogCalib. As baseline methods, we select (1) Vanilla calibration, which uniformly applies calibration loss to all data. (2) Random calibration, which randomly distinguishes between known and unknown data while maintaining a consistent number of known samples.

\begin{figure}[t]
    \centering
    \includegraphics[width=\linewidth]{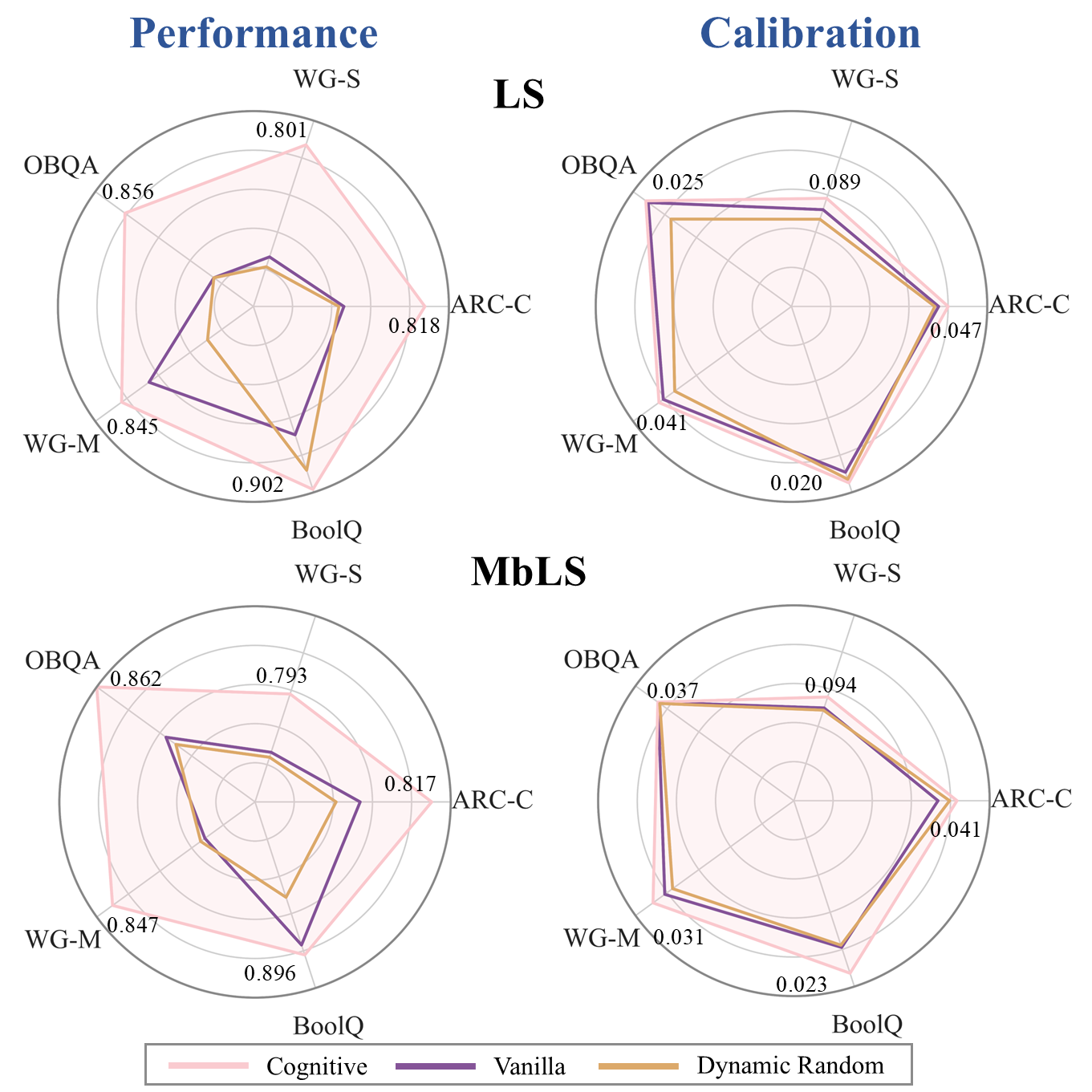}
    \caption{Comparison of CogCalib against baselines (Vanilla and Dynamic random) in terms of fine-tuning performance and calibration. Since a lower ECE is better, we normalize ECE to $[0,1]$ using $\frac{ECE_{max}-ECE}{ECE_{max}-ECE_{min}}$ ($ECE_{max}=0.2$, $ECE_{min}=0.01$). ECP's results are shown in Figure ~\ref{fig:ablation_1_ecp} of Appendix~\ref{appendix:vanilla_vs_random}.}
    \label{fig:ablation_1}
\end{figure}

As shown in Figure ~\ref{fig:ablation_1}, in these tasks, our cognitive methods achieved optimal results in both fine-tuning performance and calibration, thereby validating the necessity of employing distinct learning strategies within CogCalib. Whether using Vanilla Calibration or Random Calibration, the accuracy of downstream tasks declined (see more results in Appendix~\ref{appendix:vanilla_vs_random}). Further research revealed that applying calibration loss to unknown data impairs the model's performance on downstream tasks (detailed analysis is presented in the Appendix~\ref{appendix:effects}), namely that unknown data are critical for aligning the model with downstream tasks.
\begin{figure}[t]
    \centering
    \includegraphics[width=\linewidth]{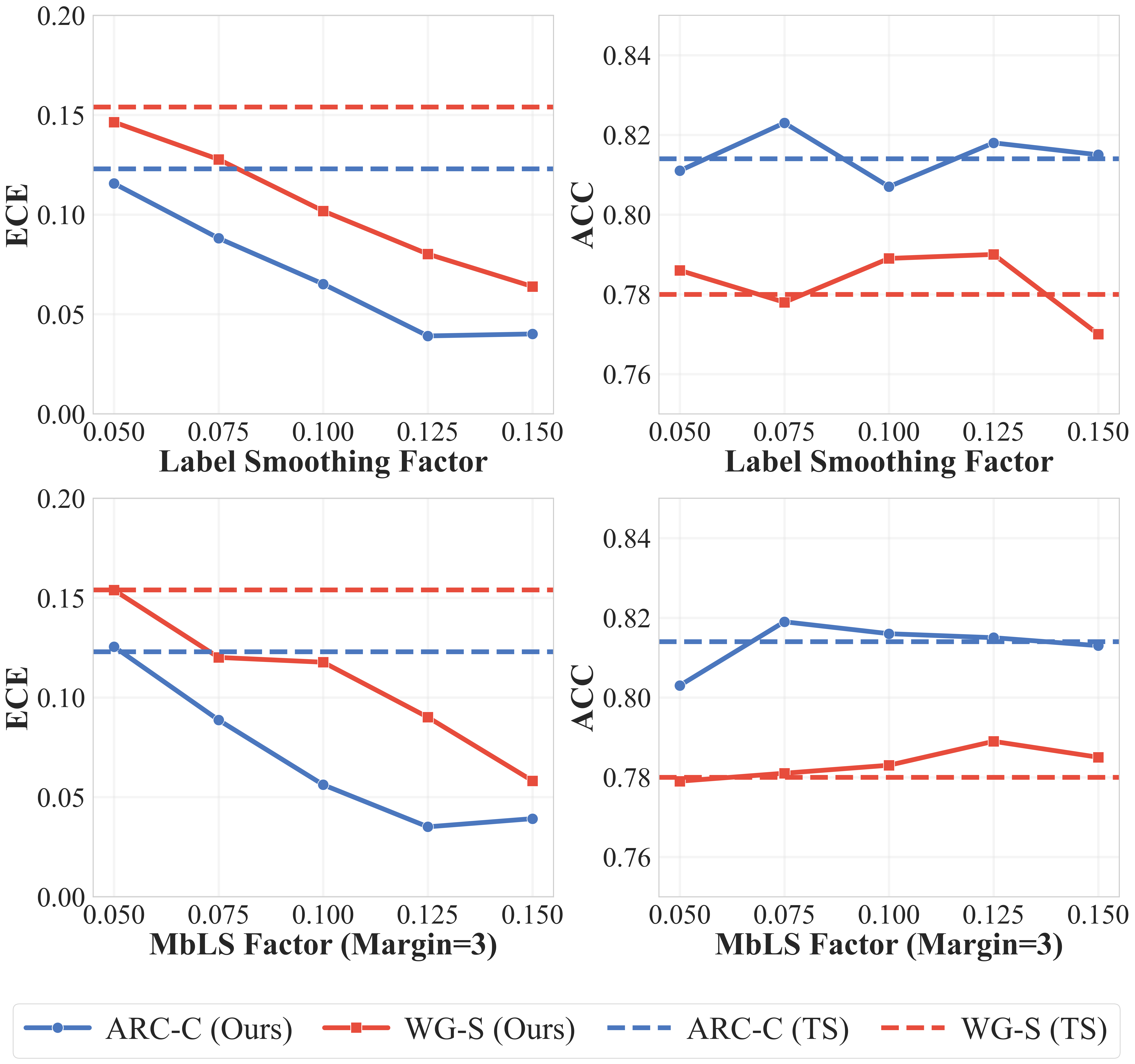}
    \caption{Sensitivity to Hyperparameters. We adjust the hyperparameters of CogCalib and compare its performance with the temperature scaling baseline on both ARC-C and WG-S datasets. The experimental results demonstrate the robustness of our method, showing consistent gains across various configurations.}
    \label{fig:abala_2}
\end{figure}

\noindent \textbf{Sensitivity to Hyperparameters} We investigate the impact of hyperparameter choices for CogCalib on performance. As illustrated in Figure~\ref{fig:abala_2}, both LS and MbLS consistently demonstrate lower ECE across various hyperparameter configurations compared to the temperature scaling baseline, while maintaining comparable accuracy. More results regarding ECP are provided in the Appendix~\ref{appendix:sens_hyper}. These findings demonstrate the robustness of CogCalib, with multiple hyperparameter configurations yielding calibration improvements.

\section{Conclusion}
In this work, we reveal that LLMs' prior knowledge causes potential poor calibration due to the ubiquitous presence of known data in real-world fine-tuning, which we discover would induce overconfidence. To address this, we propose CogCalib, a real-time cognition-aware calibration, which could achieve more objective fine-tuning. Through extensive experiments, we demonstrate that CogCalib effectively improves calibration while maintaining model performance without additional computational overhead during deployment, enabling more objective and trustworthy fine-tuning in safety-critical applications.

\section*{Limitations}
Our research focuses on how prior knowledge in large language models (LLMs) leads to poor calibration during the fine-tuning process, and we propose a real-time calibration framework to address this issue. However, our study has only investigated models up to 13B parameters, and larger-scale models remain unexplored. Given additional GPU resources, we can conduct more comprehensive experiments to validate our findings on larger models. Furthermore, our framework incorporates some calibration terms, and new calibration terms may potentially achieve better performance in the future. Nevertheless, our research provides a novel perspective on the problem of poor calibration during fine-tuning and offers a real-time solution.

\section*{Ethics Statement}
In this work, ethical considerations have been carefully addressed, and all research activities were conducted in strict compliance with the ACL Ethics Guidelines. The primary focus of this study is to investigate the impact of LLM prior knowledge on calibration while proposing a real-time calibration framework. All models and datasets utilized in this research are publicly available and have been widely adopted by the research community. The experimental results presented herein have been rigorously validated for accuracy and reproducibility. Based on these considerations, we assert that this research does not raise any ethical concerns.

\section*{Acknowledgements}

The work is supported by the grants from the Natural Science Foundation of China (62225202, 62202029), and Young Elite Scientists Sponsorship Program by CAST (No. 2023QNRC001). We owe sincere thanks to all authors for their valuable efforts and contributions. The corresponding author is Haoyi Zhou.

\bibliography{custom}

\appendix
\label{sec:Appendix}
\clearpage
\clearpage

\section{Details of Datasets and Calibration Set}
\label{Appendix:dataset}

\begin{table}[H]
\centering
\small
\begin{tabular}{ccccc}
\toprule 
\textbf{Dataset} & \textbf{Train} & \textbf{Test} & \textbf{Val} &\textbf{Calibration}\\
\midrule
\textbf{HotpotQA} & 16k & 2k & 1k & 1k \\
\textbf{MedMCQA} & 10k & 2k &  1k & 1k \\
\textbf{OBQA} & 4452  & 500 & 500 & 500\\
\textbf{BoolQ} & 8427 & 3270 & 1k & 1k\\
\textbf{ARC-C} & 1119 & 1172 & 299 & 200 \\
\textbf{WG-S} & 580 & 1267 & 80 & 80 \\
\textbf{WG-M} & 2258 & 1267 & 300 & 300\\
\bottomrule
\end{tabular}
\caption{Configuration of datasets for fine-tuning. The validation set is utilized for Temperature Scaling to search for optimal temperature for calibration, while the calibration set is employed for threshold updating during finetuning.}
\label{tab:datasets}
\end{table}
We present detailed statistics of the finetuning tasks in Table~\ref{tab:datasets}. For test-only tasks, including MMLU subtasks (Business, Culture, History, Psychology, Physics, Economics, Health, and Law) and the ARC-E task, we strictly adhered to their official dataset configurations. For datasets that originally lacked validation sets in Table~\ref{tab:datasets}, we partitioned a portion of their training data to create validation sets specifically for Temperature Scaling. The calibration set was designed to have a comparable size to the validation set, with samples randomly selected from the training set at fixed intervals for threshold updates. Notably, MedMCQA~\citep{pmlr-v174-pal22a}, a comprehensive medical multiple-choice dataset, was restructured into an open-ended format where option texts were directly used as answers, following the same question-answering format as HotpotQA.

\section{Addendum to Section 3}
\subsection{Construction of Datasets with Varying Knowledge Bias}
\label{appendix:data_construct}
In Section~\ref{sec:3}, we simulate varying knowledge bias by adjusting the ratio of unknown to known samples in the fine-tuning set. Specifically, $N_k$ and $N_{unk}$ denote the number of known and unknown samples in the original dataset, and $N = min\{N_k, N_{unk}\}$ represents the total data volume. 

When $N_{k}\leq N_{unk}$, we first form $D_{0:r}$ by including all known samples (where the subscript indicates the ratio of unknown to known data, $r=5$ in our experiments). We then construct the dataset $\\{D _ {i:(r-i)}\\} _ {i=1}^{r}$ according to these rules: randomly remove $\frac{N}{r}$ known samples from $D _ {{i}:{(r-i)}}$ and add $\frac{N}{r}$ randomly selected unknown samples from the original dataset to form $D _ {{(i+1)}:{(r-i-1)}}$. For the case where $N _ {k}>N _ {unk}$, the process follows the same principle.

\subsection{Additional Results of Section 3}
\label{appendix:more_of_sec3}
\begin{figure}
    \centering
    \includegraphics[width=\linewidth]{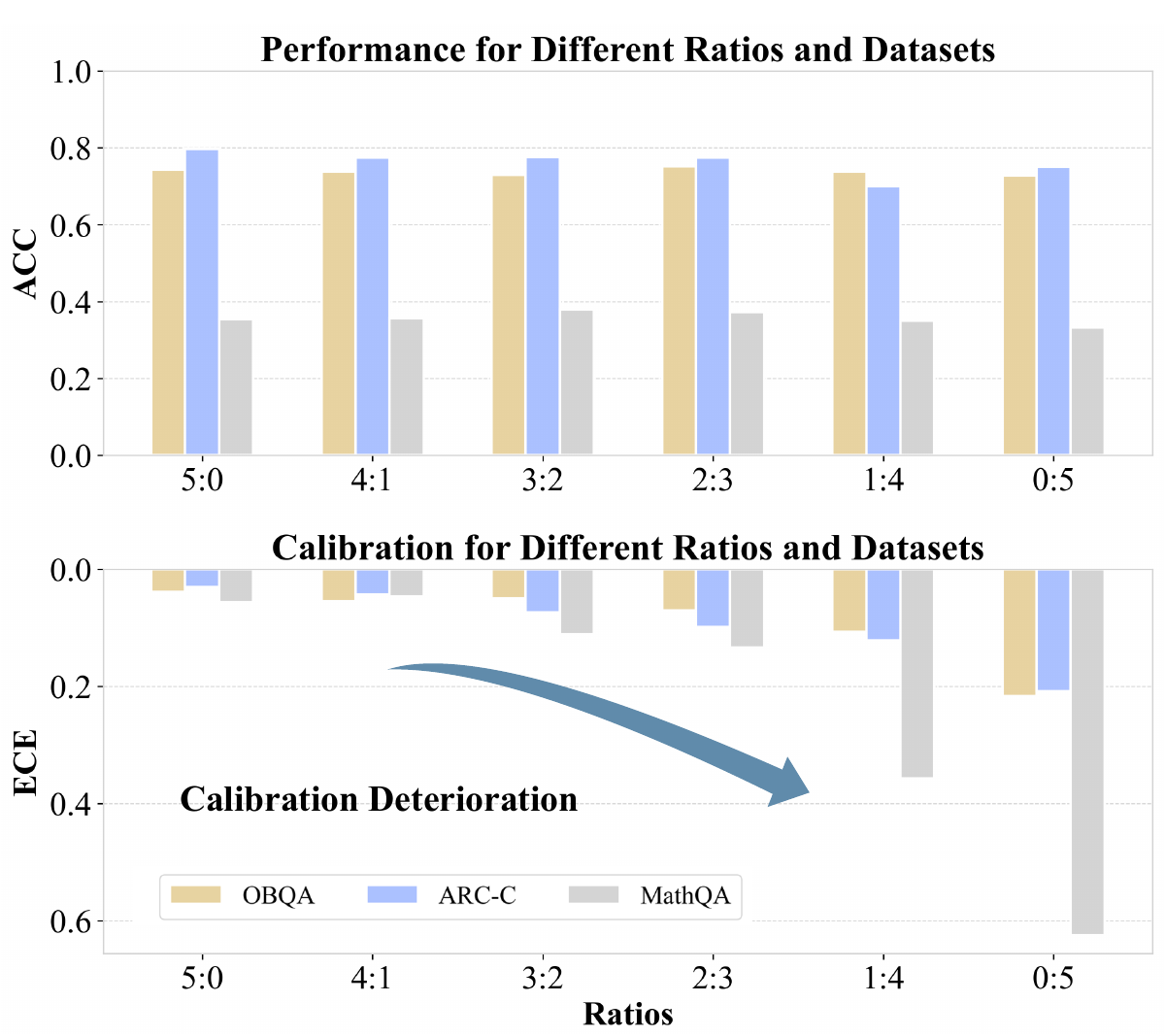}
    \caption{Accuracy and ECE of Llama3-8B fine-tuned with different knowledge biases in MathQA. The ratio varies from 5:0 to 0:5 (unknown data:known data), with equal dataset sizes. Calibration deteriorates as the knowledge bias lowers, while higher knowledge bias helps improve calibration aligning with findings in Section~\ref{sec:3.1}.}
    \label{fig:fenbu_more1}
\end{figure}

\begin{figure*}
    \centering
    \includegraphics[width=\linewidth]{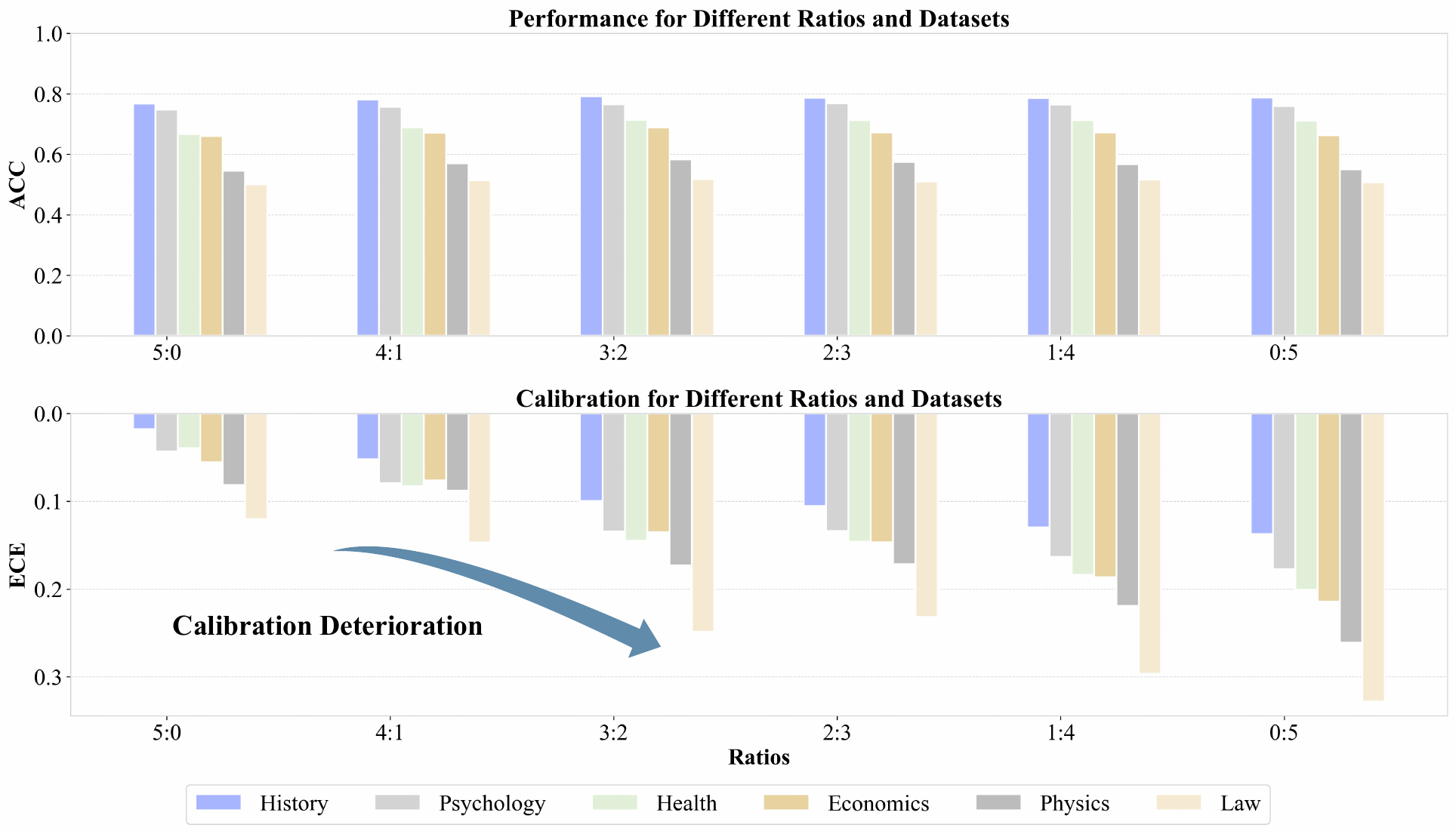}
    \caption{Accuracy and ECE of Llama3-8B fine-tuned with different knowledge biases in the open-ended dataset MedMCQA restructured by us. The ratio varies from 5:0 to 0:5 (unknown data:known data), with equal dataset sizes. Calibration also deteriorates as the knowledge bias lowers, while higher knowledge bias helps improve calibration in open-ended fine-tuning scenarios, aligning with findings in Section~\ref{sec:3.1}.}
    \label{fig:fenbu_more2}
\end{figure*}

In Section~\ref{sec:3.1}, we demonstrated that low-bias leads to overconfidence, while high-bias data contributes to better calibration. This section presents additional experimental evidence supporting this phenomenon across multiple datasets, following the experimental protocol established in Section~\ref{sec:3.1}. Figure~\ref{fig:fenbu_more1} illustrates the ECE metrics and fine-tuning performance obtained from experiments on MathQA, where we constructed scenarios with varying degrees of knowledge bias. Figure~\ref{fig:fenbu_more2} presents our test results on MedMCQA, a domain-specific open-ended dataset which is restructured by us as explained in Appendix~\ref{Appendix:dataset}. Both figures clearly demonstrate that calibration performance deteriorates significantly as bias decreases. These findings further corroborate our conclusion from Section~\ref{sec:3.1}, supporting the principle of "minimal bias, maximal overconfidence". Furthermore, additional experiments are conducted on Llama2-13B (Figure \ref{fig:fenbu_llama2}), Qwen2.5-7B (Figure \ref{fig:fenbu_qwen}), and Mistral-7B (Figure \ref{fig:fenbu_mistral}), following the same setup described in Figure \ref{fig:distribution_impact}. The results substantiate that calibration degradation is a consistent phenomenon observed across various LLMs, regardless of their architecture or parameter scale.

\begin{figure}
    \centering
    \includegraphics[width=\linewidth]{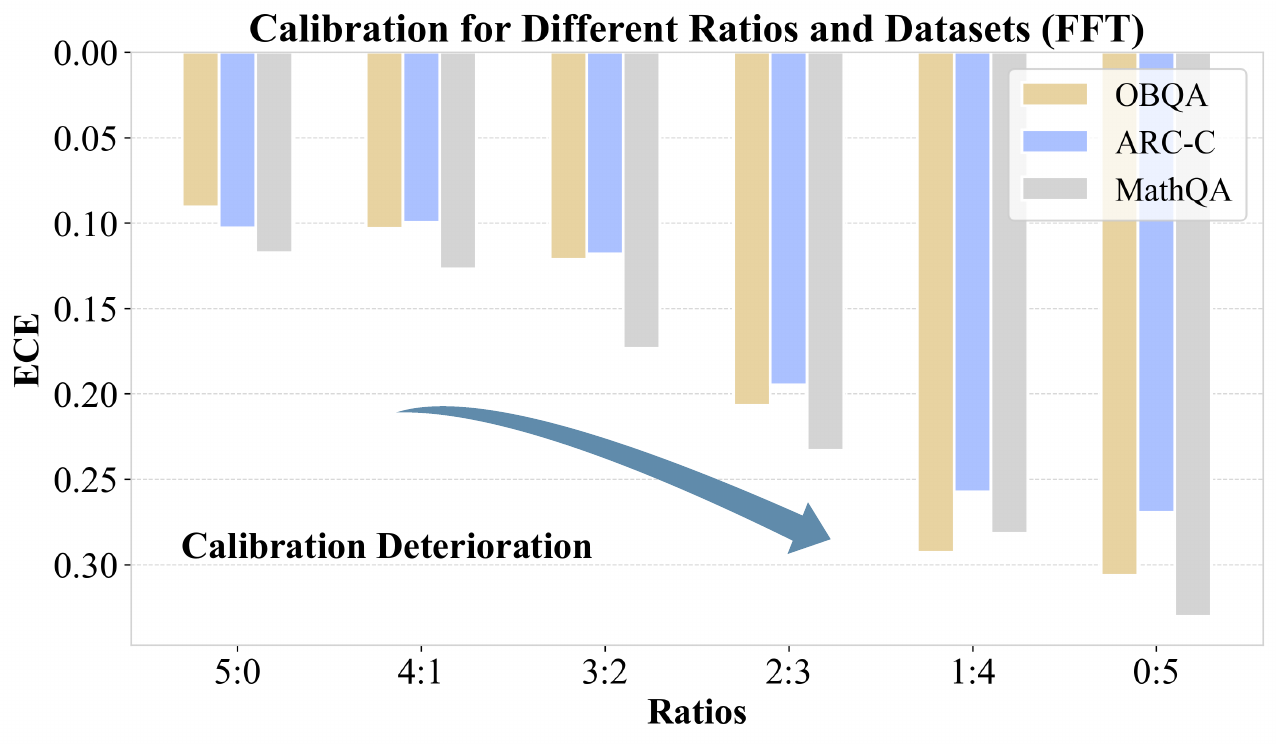}
    \caption{ECE of Llama3-8B fine-tuned with different knowledge biases in MathQA using Full Fine-Tuning (FFT). The ratio varies from 5:0 to 0:5 (unknown data:known data), with equal dataset sizes. Calibration deteriorates as the knowledge bias lowers, while higher knowledge bias helps improve calibration aligning with findings in Section~\ref{sec:3.1}.}
    \label{fig:fenbu_fft_mathqa}
\end{figure}

\begin{figure}
    \centering
    \includegraphics[width=\linewidth]{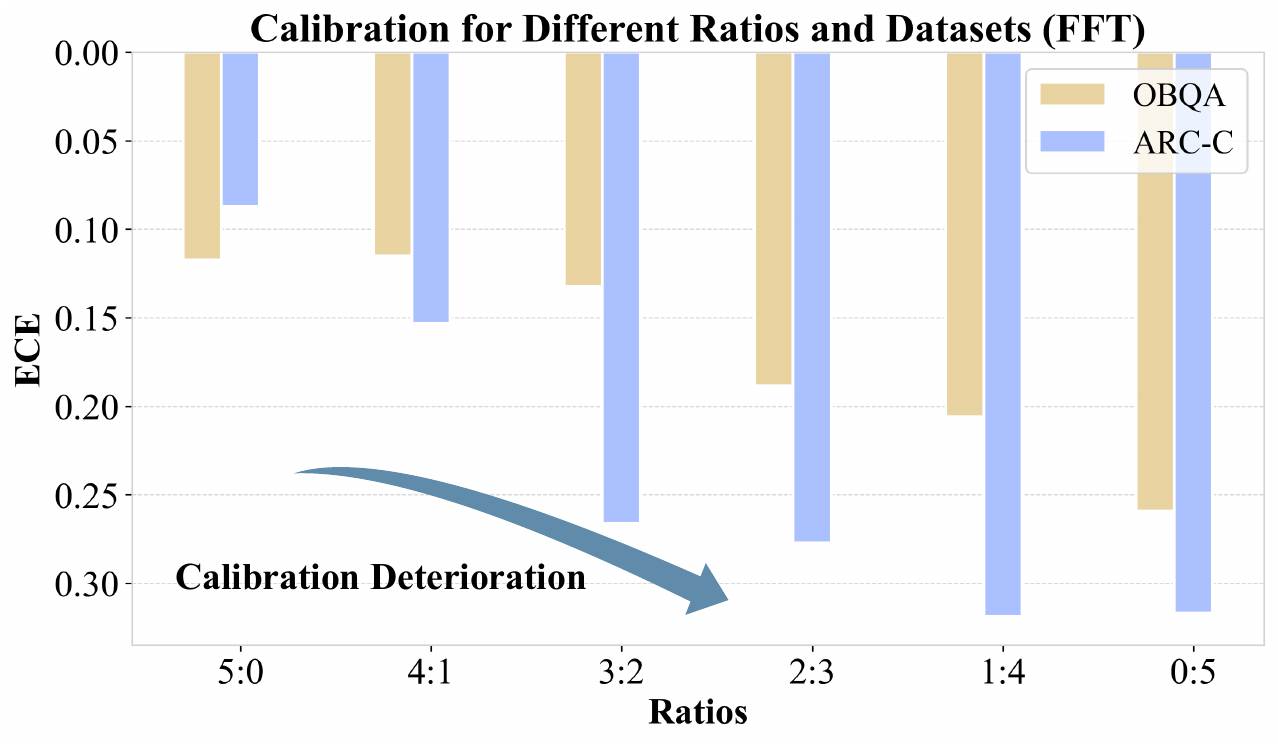}
    \caption{ECE of Llama3-8B fine-tuned with different knowledge biases in ARC-C using Full Fine-Tuning (FFT). The ratio varies from 5:0 to 0:5 (unknown data:known data), with equal dataset sizes. Calibration deteriorates as the knowledge bias lowers, while higher knowledge bias helps improve calibration aligning with findings in Section~\ref{sec:3.1}.}
    \label{fig:fenbu_fft_arcc}
\end{figure}

\begin{figure}
    \centering
    \includegraphics[width=\linewidth]{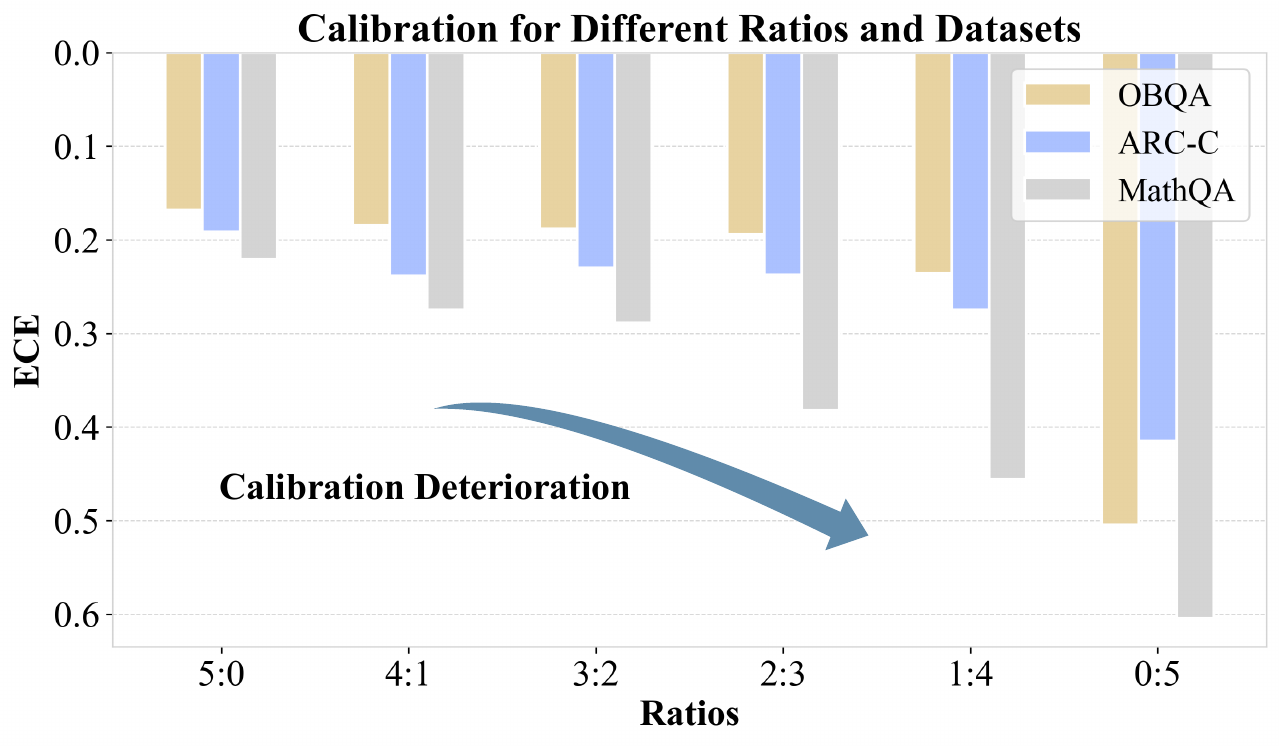}
    \caption{ECE of Llama2-13B fine-tuned with different knowledge biases following the same setup as Figure \ref{fig:distribution_impact}. The results confirm that calibration degradation is a consistent phenomenon across models.}
    \label{fig:fenbu_llama2}
\end{figure}

\begin{figure}
    \centering
    \includegraphics[width=\linewidth]{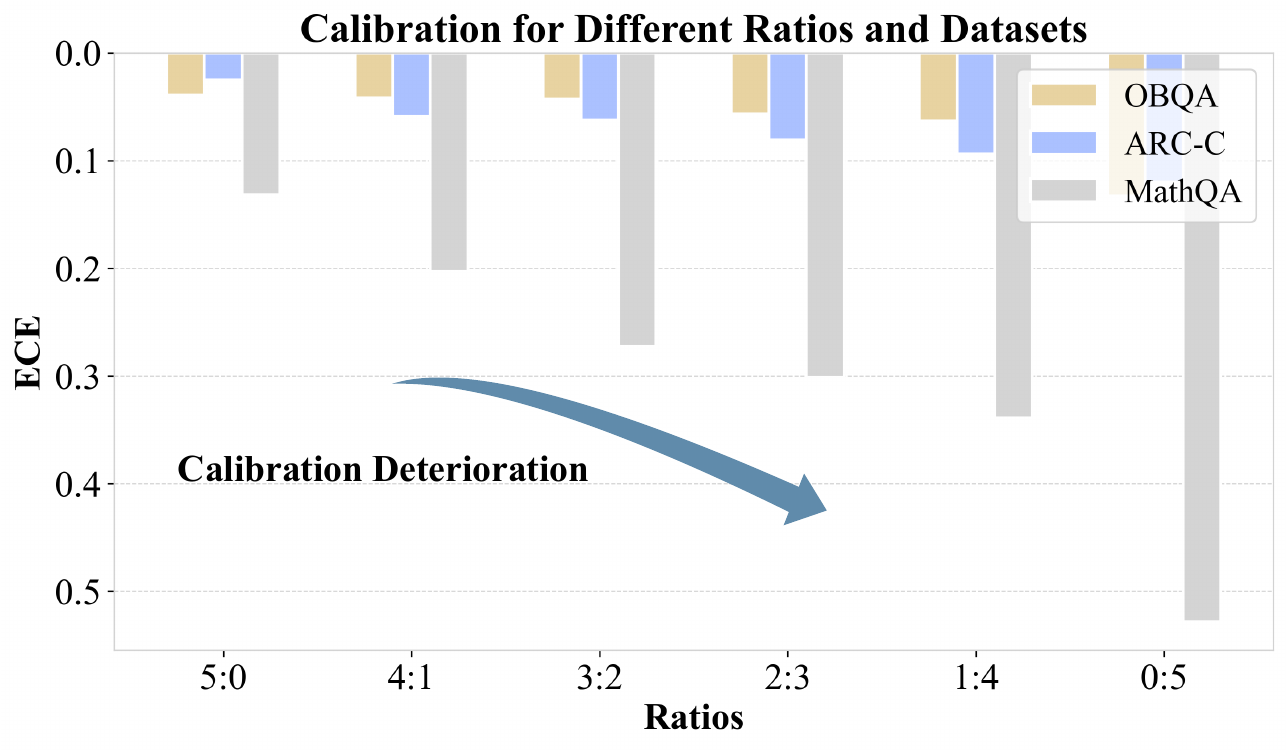}
    \caption{ECE of Qwen2.5-7B fine-tuned with different knowledge biases following the same setup as Figure \ref{fig:distribution_impact}. The results confirm that calibration degradation is a consistent phenomenon across models.}
    \label{fig:fenbu_qwen}
\end{figure}

\begin{figure}
    \centering
    \includegraphics[width=\linewidth]{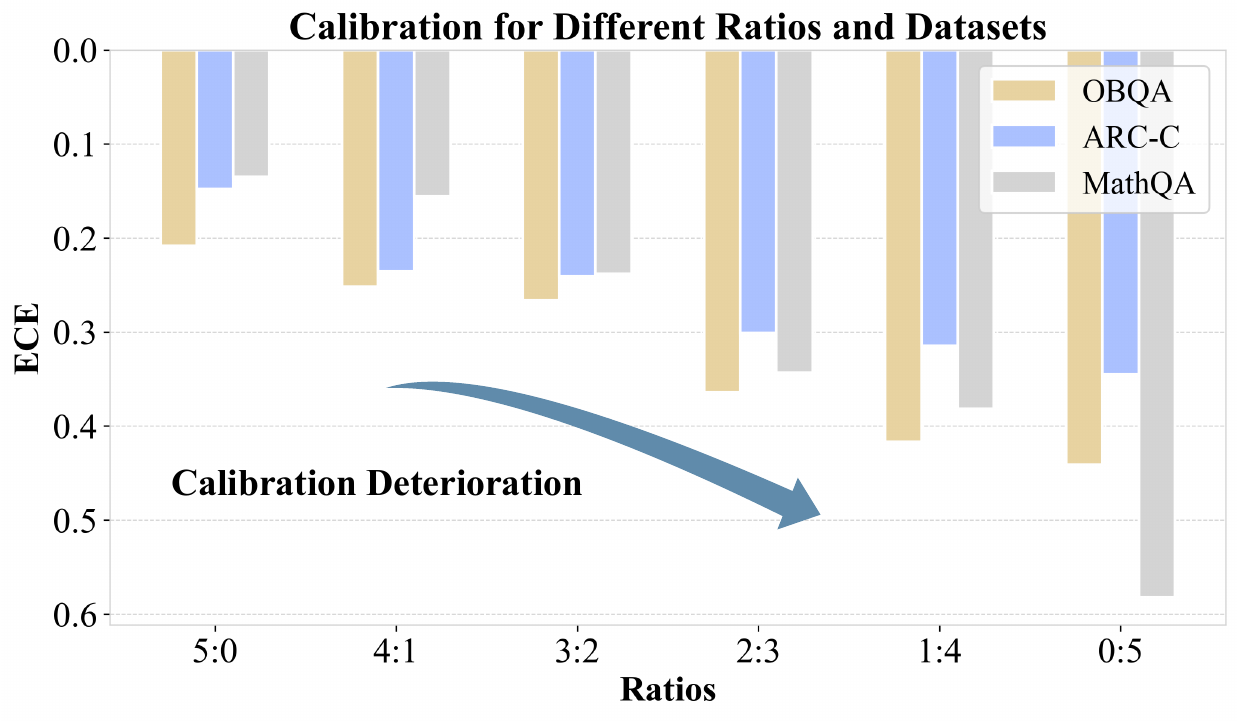}
    \caption{ECE of Mistral-7B fine-tuned with different knowledge biases following the same setup as Figure \ref{fig:distribution_impact}. The results confirm that calibration degradation is a consistent phenomenon across models.}
    \label{fig:fenbu_mistral}
\end{figure}

To examine whether this phenomenon extends to Full Fine-Tuning (FFT) scenarios, we conducted additional experiments using Llama3-8B model on both MathQA and ARC-C datasets. The results, visualized in Figure~\ref{fig:fenbu_fft_mathqa} and Figure~\ref{fig:fenbu_fft_arcc}, reveal that the pattern persists in FFT settings. Low-bias data consistently leads to model overfitting, while the introduction of new knowledge helps mitigate this effect and improves calibration. The observation of this pattern in FFT scenarios further strengthens our findings, suggesting that the relationship between bias and calibration is a robust phenomenon that transcends specific training approaches.

As a supplement, to ensure the stability and consistency of the experiments in Figure \ref{fig:verify-id-conf}b, standard deviations from three trials with different random seeds are presented in Table \ref{tab:fig3b_std}. These values demonstrate the consistency and reliability of the results.

\begin{figure}[!htbp]
    \centering
    \includegraphics[width=\linewidth]{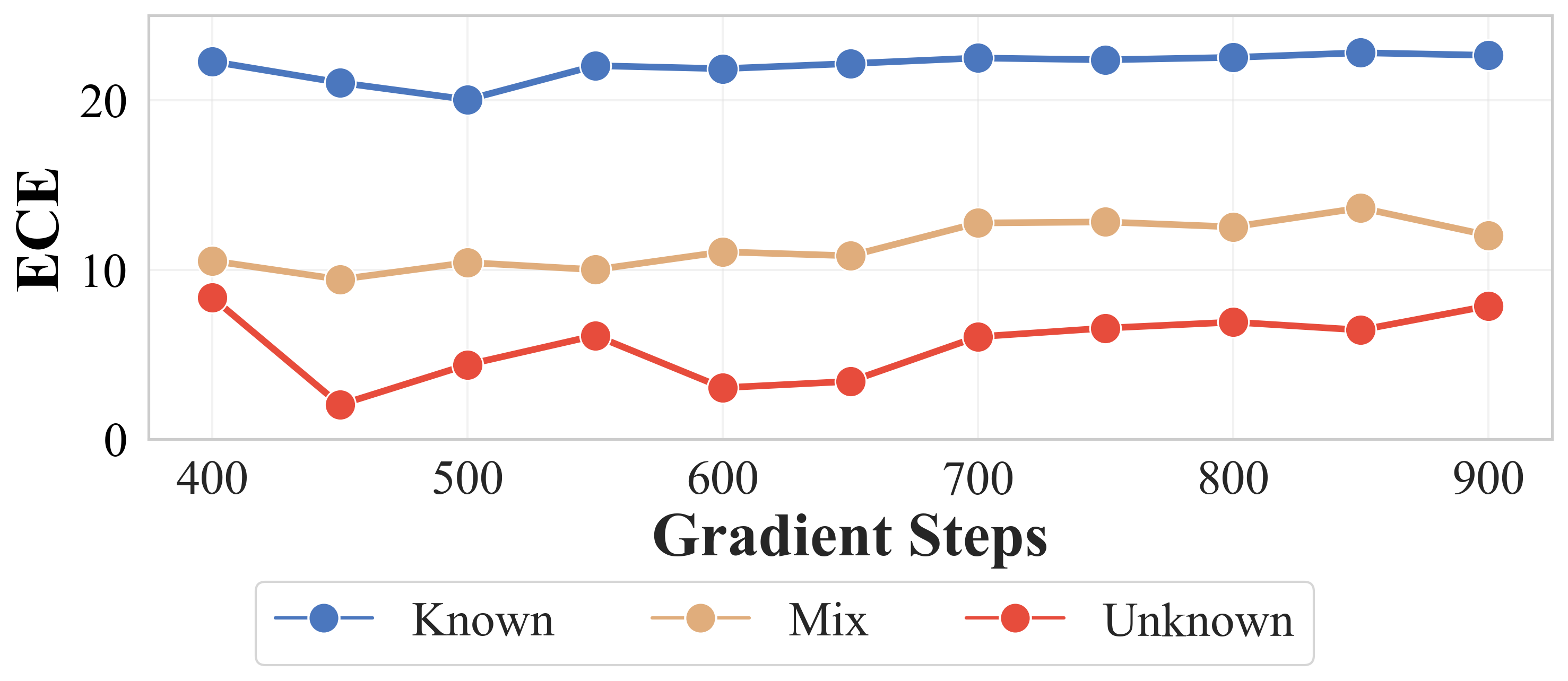}
    \caption{ECE of Llama3-8B after fine-tuning on unknown (high-bias), mixed, and known (low-bias) datasets, where the mixed dataset is randomly sampled from OBQA with an equal size. ECE curve for models fine-tuned on the mixed dataset maintains an intermediate position, indicating low-bias data would dilute the calibration benefits of high-bias data.}
    \label{fig:verify-ece}
\end{figure}

Additionally, to investigate the potential interactions between different knowledge bias levels across datasets in realistic scenarios, we conducted a comparative analysis using the OBQA dataset. We randomly sampled equal portions of pure known data (low bias), pure unknown data (high bias), and mixed data. Figure~\ref{fig:verify-ece} illustrates the calibration results after fine-tuning the model on these 3 dataset categories. The ECE curve for models fine-tuned on the mixed dataset consistently maintains an intermediate position between the other two curves throughout the fine-tuning process. This observation suggests that low-bias data effectively dilutes the calibration benefits achieved through high-bias fine-tuning.
\begin{table*}
\centering
\small
\begin{tabular}{lcccccccccccc}
\toprule
Steps                & 50   & 100  & 150  & 200  & 250  & 300  & 350  & 400  & 450  & 500  & 550  & 600  \\ 
\midrule
Conf of CS (known)   & 0.34 & 1.16 & 0.71 & 1.34 & 0.66 & 1.05 & 0.71 & 0.88 & 0.45 & 0.65 & 0.36 & 0.66 \\ 
Conf of IS (known)   & 1.76 & 3.24 & 0.17 & 3.15 & 3.47 & 3.11 & 1.33 & 0.58 & 1.10 & 0.88 & 1.73 & 1.76 \\ 
Conf of CS (unknown) & 5.48 & 6.67 & 5.26 & 2.89 & 5.01 & 0.51 & 3.01 & 4.97 & 2.88 & 4.24 & 2.87 & 1.82 \\ 
Conf of IS (unknown) & 3.66 & 5.61 & 3.03 & 2.29 & 3.76 & 0.63 & 3.57 & 4.24 & 4.65 & 4.51 & 3.72 & 1.19 \\ 
\bottomrule
\end{tabular}
\caption{Standard deviation of confidence values at different steps. Standard deviations are recorded from three trials with different random seeds. These values demonstrate the consistency and reliability of the results.}
\label{tab:fig3b_std}
\end{table*}

\section{Details of LS, MbLS and ECP}
\label{Appendix:cal_methods}
\subsection{Label Smoothing}
Label Smoothing (LS)~\citep{szegedy2016rethinking} replaces one-hot encoded labels with smoothed distributions by allocating small probabilities to non-target classes, effectively reducing model overconfidence and improving calibration performance in deep neural networks, as follows:
\begin{equation}
\mathcal{L}_{\mathrm{I.S}}=-\sum_{k}((1-\epsilon)q_{k}+\frac{\epsilon}{K})\log p_{k},
\end{equation}
where $\epsilon$ is label smoothing factor, $K$ denotes the numbers of total classes, $\mathbf{p}$ is the softmax probability predictions by model, which is computed as follows:
\begin{equation}
\mathbf{p}=(p_k)_{1\leq k\leq K}\in\mathbb{R}^K;\quad p_k=\frac{\exp^{l_k}}{\sum_j^K\exp^{l_j}},
\end{equation}
where $\mathbf{l}=(l_k)_{1\leq k\leq K}\in\mathbb{R}^K$ denotes logits vectors. Furthermore, the loss function of LS can be written as~\citep{liu2022devil}:
\begin{equation}
    \mathcal{L}_{\mathrm{LS}}\triangleq\mathcal{L}_{\mathrm{CE}}+\frac{\epsilon}{1-\epsilon}\mathcal{D}_{\mathrm{KL}}\left(\mathbf{u}||\mathbf{p}\right),    
\end{equation}
where $L_{CE}$ denotes Cross-Entropy loss, $\mathbf{u}$ denotes uniform distribution $\mathbf{u}=\frac{1}{K}$, $\triangleq$ stands for equality up to additive and/or nonnegative multiplicative constants.

\subsection{Margin-based Label Smoothing}
Margin-based Label Smoothing (MbLS)~\citep{liu2022devil} addresses the calibration issue in deep neural networks by imposing inequality constraints on logit distances, unlike traditional methods that use equality constraints. This approach provides a better balance between model discrimination and calibration performance. MbLS introduces inequality constraints with controllable margins as follows:
\begin{equation}
\mathcal{L}_{MbLS}=\mathcal{L}_{\mathrm{CE}}+\gamma\sum_k\max(0,\max_j(l_j)-l_k-m),
\end{equation}
where $\gamma$ is the label smoothing factor, and $m$ is the logits margin.

\subsection{ECP}
ECP ~\citep{pereyra2017regularizing} is a neural network regularization technique that works by penalizing low entropy output distributions as follows:
\begin{equation}
    \mathcal{L}_{ECP}=\mathcal{L}_\mathrm{CE}-\beta\mathcal{H}(\mathbf{p}),
\end{equation}
where $\beta$ is ECP factor, and $\mathcal{H}$ denotes the Shannon entropy of the softmax prediction given by: 
\begin{equation}
    \mathcal{H}(\mathrm{p})=-\sum_kp_k\log(p_k).
\end{equation}

\section{Details of Style Adaptation}
\label{appendix:adaptation}
As shown in Table~\ref{tab:abalation_warm}, the AUROC scores for $t_0$ discrimination demonstrate significant improvement with style adaptation. This improvement can be attributed to better alignment between the model's output style and fine-tuning data patterns after style adaptation, which enhances the model's capability to utilize NLL for distinguishing between known and unknown samples.

\begin{table}[H]
\centering
\small
\begin{tabular}{ccc}
\toprule 
\textbf{Dataset} & \textbf{$t_0$ w/o sa} & \textbf{$t_0$ w/ sa} \\
\midrule
\textbf{HotpotQA} & 0.729 & 0.888 \\
\textbf{MedMCQA} & 0.712 & 0.908 \\
\bottomrule
\end{tabular}
\caption{AUROC scores for $t_0$ discrimination in Llama3-8B with/without style adaptation (sa) on Open-End tasks.}
\label{tab:abalation_warm}
\end{table}

\section{Details of ECE and Reliability Diagram}
\label{appendix:ece}
Expected Calibration Error (ECE) serves as one of the primary metrics for assessing calibration, measuring the alignment between model confidence and accuracy. As demonstrated in Equation (\ref{eq:ece}), ECE operates by partitioning model confidence (maximum output probabilities) into $m$ bins, then computing a weighted sum of the discrepancies between accuracy and confidence across all bins. 
\begin{equation}
\label{eq:ece}
ECE=\sum_{m=1}^M \frac{|B_m|}{N}|acc(B_m)-conf(B_m)|,
\end{equation}
where $|Bm|$ represents the number of samples in bin m, $N$ denotes the total number of samples, while $acc(B_m)$ and $conf(Bm)$ are the average accuracy and average confidence in bin m, respectively.

\begin{figure*}[htbp]
  \centering
  \subfloat[known, OBQA, ECE=0.134]
  {\includegraphics[width=0.33\textwidth]{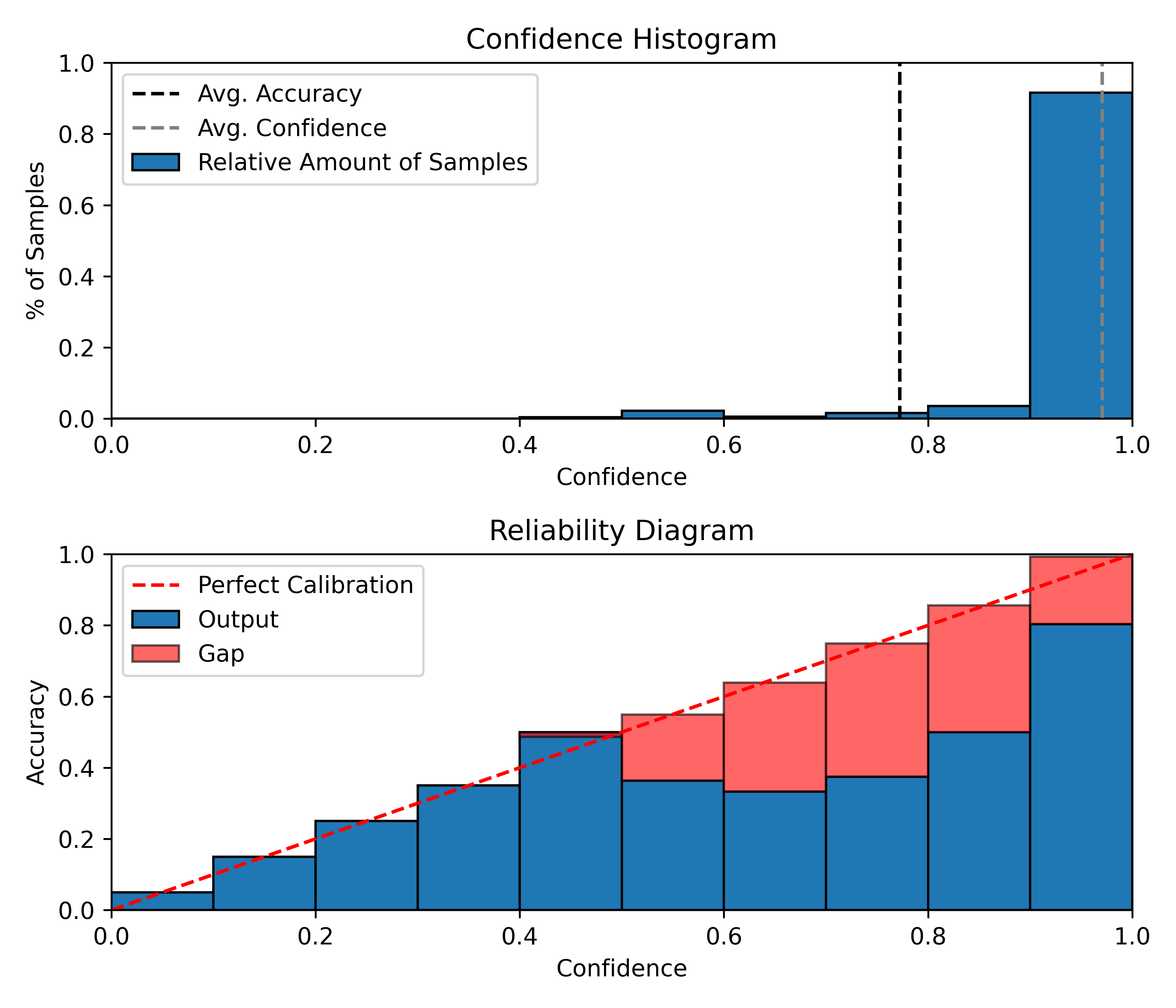}\label{fig:subfig1}}     
  \subfloat[known, ARC-C, ECE=0.148]
  {\includegraphics[width=0.33\textwidth]{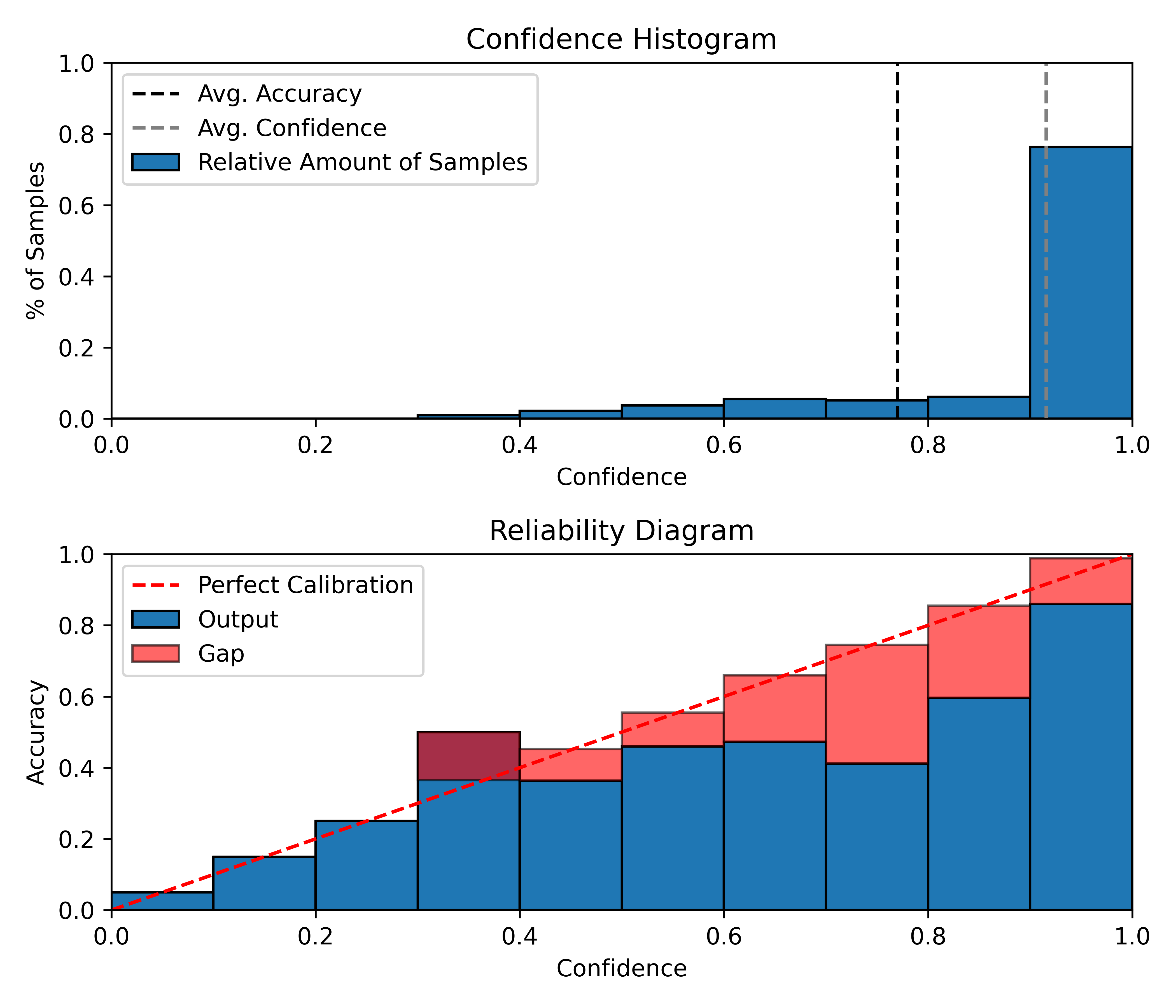}\label{fig:subfig2}}
  \subfloat[known, MathQA, ECE=0.330]
  {\includegraphics[width=0.33\textwidth]{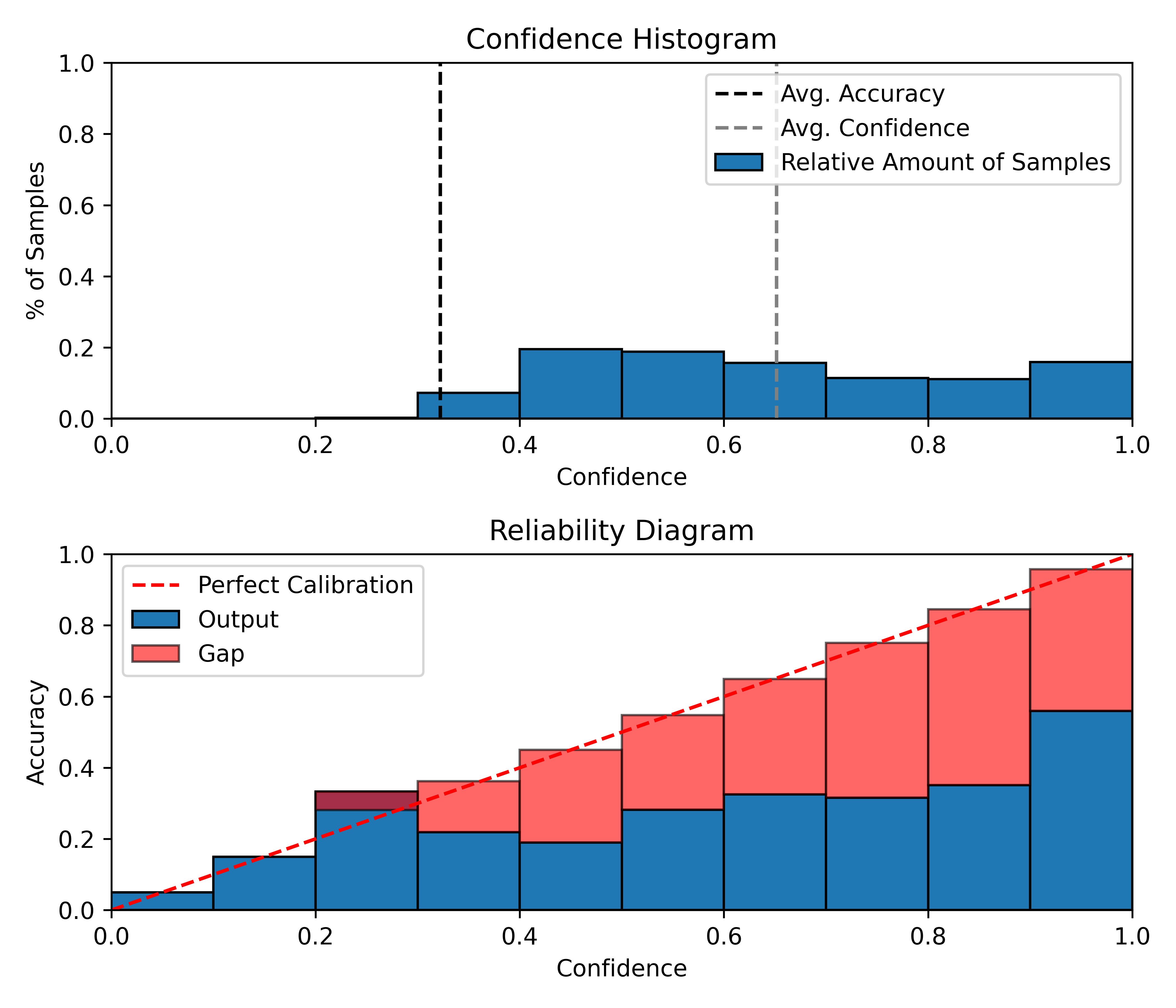}\label{fig:subfig3}} \\
  \subfloat[unknown, OBQA, ECE=0.047]
  {\includegraphics[width=0.33\textwidth]{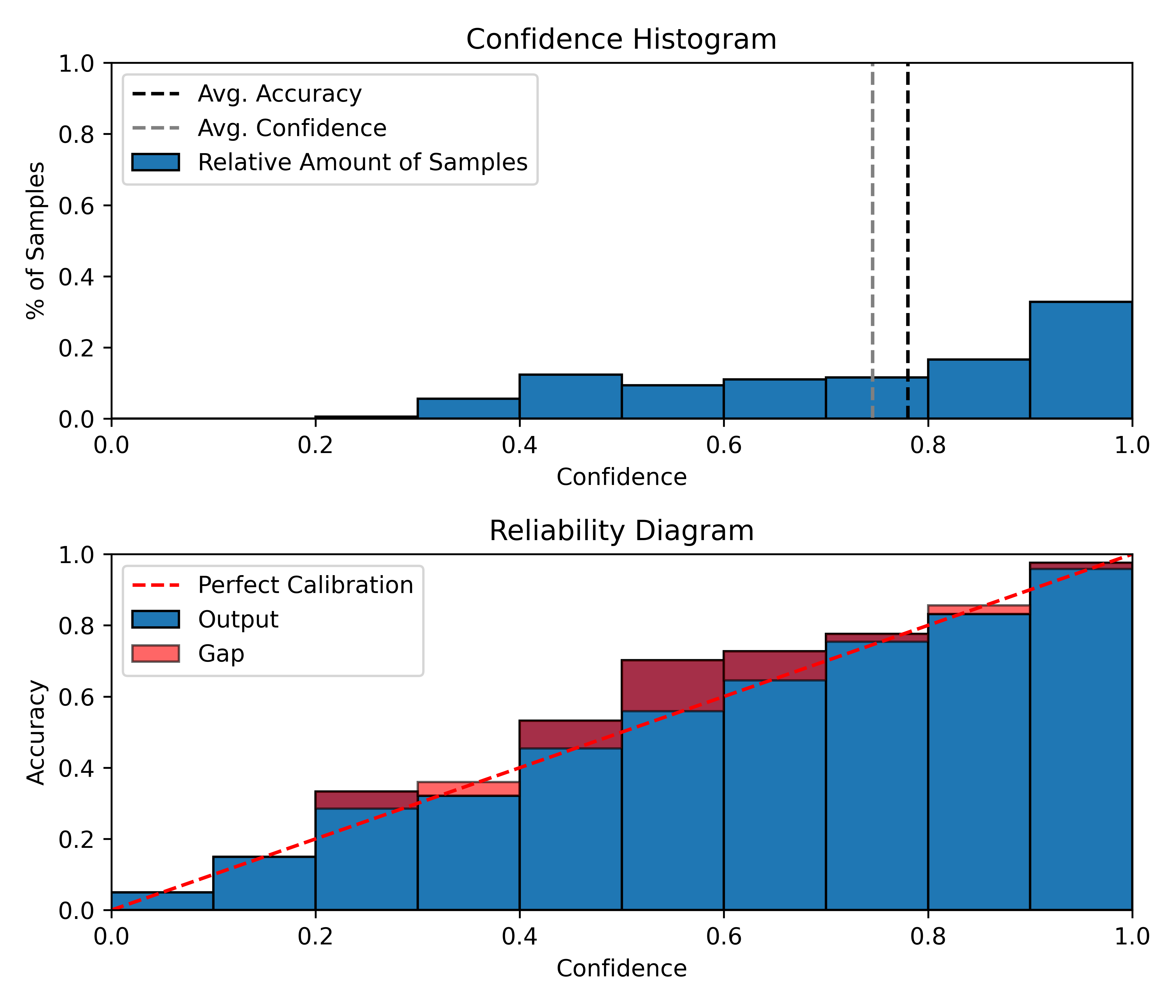}\label{fig:subfig4}}
  \subfloat[unknown, ARC-C, ECE=0.032]
  {\includegraphics[width=0.33\textwidth]{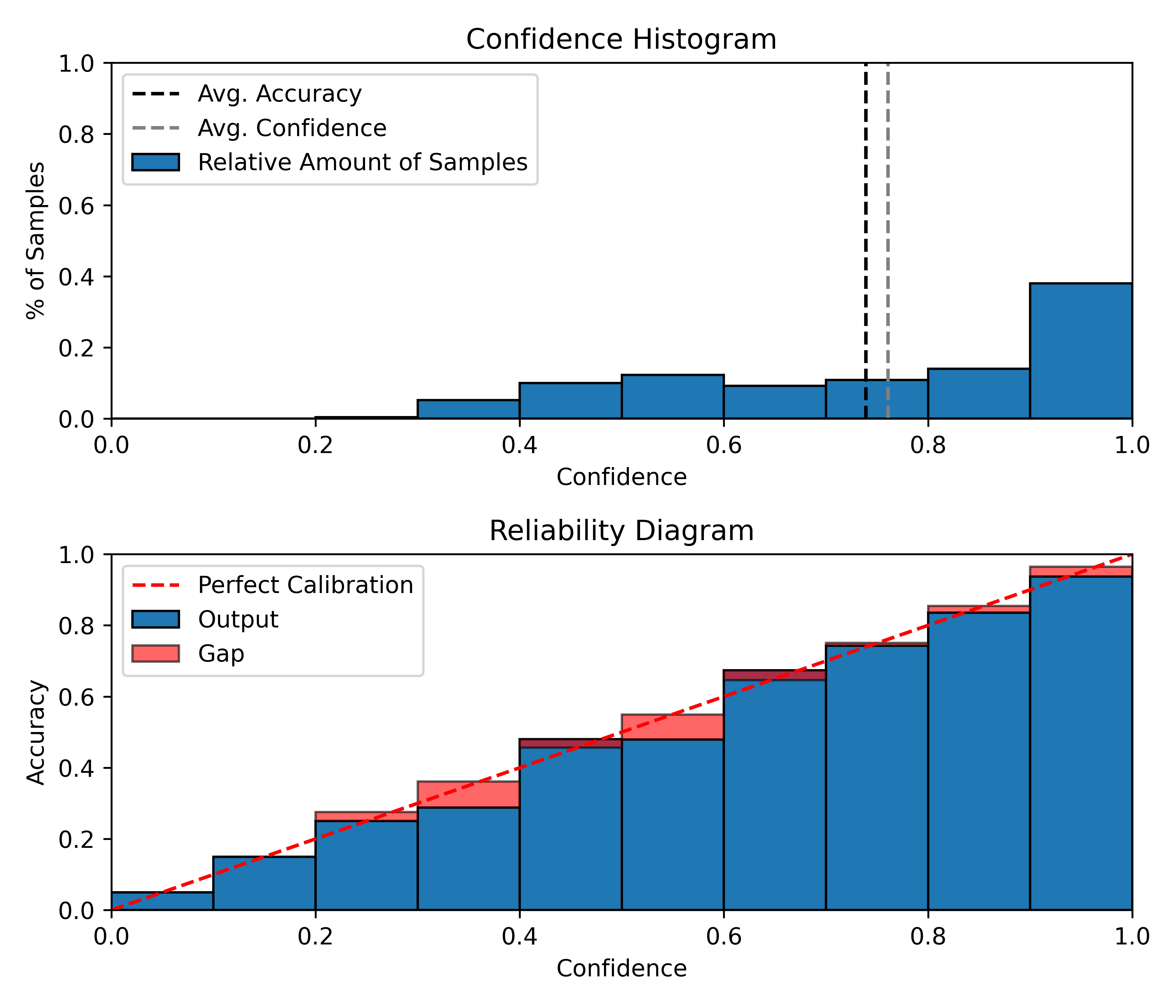}\label{fig:subfig5}}
  \subfloat[unknown, MathQA, ECE=0.102]
  {\includegraphics[width=0.33\textwidth]{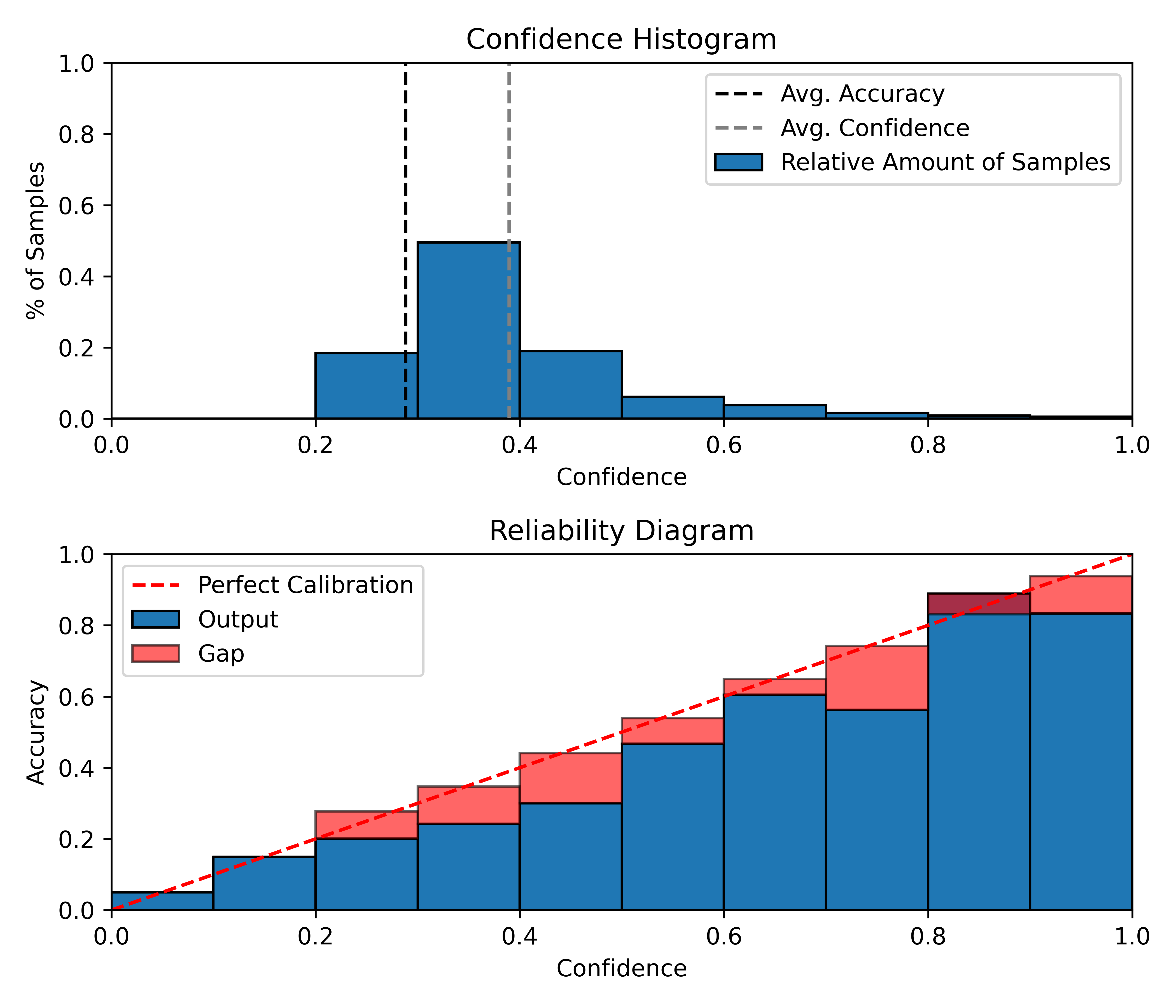}\label{fig:subfig6}}
  \caption{Reliability diagrams of models fine-tuned on OBQA known data or unknown data, evaluated on both ID test and OOD test (ARC-C, MathQA). Models trained on unknown data demonstrate better alignment between confidence and accuracy, further validating the conclusions drawn in Section~\ref{sec:3.1}.}
  \label{fig:relia}
\end{figure*}

In addition to ECE, calibration property can be visualized through Reliability Diagram~\citep{brocker2007increasing}. As illustrated in the Figure~\ref{fig:relia}, there are significant differences in the reliability diagrams between models fine-tuned with unknown data versus known data. This distinction is evident in both ID and OOD scenarios, further corroborating the findings presented in Section~\ref{sec:3}.

\section{Additional Experimental Analysis}
\label{appenix:exp}
In this section, we demonstrate the versatility of the CogCalib through extensive experiments across a broader range of models and fine-tuning approaches. Our experiments encompass models of varying architectures and sizes, along with full-parameter fine-tuning methods (FFT). Additionally, we present supplementary experiments that evaluate the potential damage of calibrating unknown data on downstream tasks performance and assess the robustness of our method under different hyperparameter settings and threshold computation. Meanwhile, we provide complementary results comparing CoECP with Vanilla and Random Calibration approaches there.
\subsection{Results of Llama3-8B on MedMCQA}
\label{appendix:llama3-8b}
We modified the question-answering format of MedMCQA and transformed it into an open-ended dataset for the medical domain, as explained in Appendix~\ref{Appendix:dataset}. Table~\ref{tab:hot_ood} presents the comprehensive experimental results of Llama3-8B on OOD datasets. The cognitive methods demonstrated superior calibration performance while maintaining comparable accuracy relative to baseline approaches. Different cognitive methods exhibited varying advantages across distinct datasets. Specifically, CoECP achieved optimal ECE on 4 datasets: Physics, Economics, Health, and Law. Meanwhile, CoMbLS showed superior ECE on both OBQA and ARC-C datasets. Notably, CoECP not only excelled in calibration but also maintained leading accuracy scores across almost all test sets. 

It's worth noting that for a fairer comparison of all methods in the OOD scenario, the temperature scaling baseline uses the best temperature found on the ID data when applied to OOD data. For long texts, we used the geometric mean of probabilities as confidences~\citep{liu2023litcab} to find the optimal calibration temperature.

\begin{table*}
\centering
\small
\begin{tabular}{cccccccc} 
		\toprule 
		\textbf{Metric} &\textbf{Methods} &Physics &Economics&Health&Law &OBQA&  ARC-C     \\
		\midrule
            \multirow{8}{*}{\textbf{ECE}$\downarrow$}&
		\textbf{Vanilla SFT}& 22.87 & 18.44 & 20.00 & 28.82 & 12.71 & 15.87\\
		& \textbf{MCD} & 18.15 &  15.44 & 15.94& 24.88 & 6.93 & 10.72\\
		& \textbf{Ensemble} & 20.43& 15.52  & 17.68 & 25.45 & 7.49 & 10.11\\
		& \textbf{TS} & 25.22 & 28.53 & 26.17 &21.39  & 40.71 & 37.89\\
		\cmidrule{2-8} 
		& \textbf{CoLS} &11.51  &  7.88 & 8.89 & 15.14 & 5.39 & 4.24\\	
		& \textbf{CoMbLS} & 18.52 &14.14 &   14.26&   22.28 & \textbf{3.83} & \textbf{2.99}\\	
        & \textbf{CoECP} & \textbf{9.14}& \textbf{6.26}  & \textbf{6.58} & \textbf{14.89} & 11.50 & 13.49\\
        \cmidrule{1-8} 

        \multirow{8}{*}{\textbf{ACC}$\uparrow$}&
		\textbf{Vanilla SFT}& 55.78 & 67.78 & 68.70 & \textbf{52.30} & 59.20 & 64.84\\
		& \textbf{MCD} & 55.46 &  65.22 &67.43 & 49.63 & 59.00 & 64.50\\
		& \textbf{Ensemble} &55.00 &  67.78 &  68.90& 50.70 & 64.00 & 69.02\\
		& \textbf{TS} & 55.78&67.78  & 68.70 & \textbf{52.30} & 59.20 & 64.84\\
		\cmidrule{2-8} 
		& \textbf{CoLS} & 54.37 & 66.30 &68.35  & 51.27 & 67.00 & 70.64\\	
		& \textbf{CoMbLS} &55.15& 66.84 & 69.08 &  50.31 & 70.60 & 74.57\\	
        & \textbf{CoECP} & \textbf{56.71} &  \textbf{68.46}& \textbf{69.69} & 50.76 & \textbf{71.00} & \textbf{75.60}\\
		\bottomrule
        
\end{tabular}
\caption{Comparison of our method's performance against baseline approaches on out-of-domain (OOD) datasets is presented. The results are evaluated on the Llama3-8B model, which is fine-tuned on the open-ended HotpotQA dataset.}
\label{tab:hot_ood}
\end{table*}
\begin{table*}
\centering
\small
\begin{tabular}{cccccccc} 
		\toprule 
		\textbf{Metric} &\textbf{Methods} &Physics &Economics&Health&Law &OBQA&  ARC-C     \\
		\midrule
            \multirow{8}{*}{\textbf{ECE}$\downarrow$}&
		\textbf{Vanilla SFT}& 18.14 & 14.07 & 16.49 & 22.55 & 8.00 & 6.87\\
		& \textbf{MCD} & 15.35 &  11.16 & 13.61 & 19.82 & 5.81 & 5.19\\
		& \textbf{Ensemble} &15.26 &  9.27 & 12.51 & 21.53 & 5.60 & 5.50\\
		& \textbf{TS} & 34.80& 40.10  & 36.80 & 34.20 & 54.80 & 58.60\\
		\cmidrule{2-8} 
		& \textbf{CoLS} & 10.39 & 6.01 & 8.84 & \textbf{14.08} & 4.96 & 2.75\\	
		& \textbf{CoMbLS} & \textbf{9.81}& \textbf{5.28} & \textbf{7.90} & 14.70 & \textbf{3.11} & 2.85\\	
        & \textbf{CoECP} & 13.39 &  7.82& 9.42 & 16.98 & 5.30 & \textbf{2.44}\\
        \cmidrule{1-8} 

        \multirow{8}{*}{\textbf{ACC}$\uparrow$}&
		\textbf{Vanilla SFT}& 51.72 & 59.16 & 59.21 & 48.21 & 65.20 & 73.55\\
		& \textbf{MCD} & 53.13 &  59.03 & 59.15& 48.44 & 65.20 & 72.27\\
		& \textbf{Ensemble} & \textbf{53.75}& \textbf{64.02}  & \textbf{61.95} & 48.67 & \textbf{67.60} & \textbf{74.49}\\
		& \textbf{TS} & 51.72 & 59.16 & 59.21 &48.21  & 65.20 & 73.55\\
		\cmidrule{2-8} 
		& \textbf{CoLS} &53.13  &  59.03 & 59.15 & 48.44 & 65.20 & 72.27\\	
		& \textbf{CoMbLS} & 51.72 &58.63 & 57.20&   47.65 & 65.00 & 72.01\\	
        & \textbf{CoECP} & 51.72& 60.51  & 60.49 & \textbf{48.72} & 65.40 & 73.46\\
		\bottomrule
        
\end{tabular}
\caption{Comparison of our method's performance against baseline approaches on out-of-domain (OOD) datasets is presented. The results are evaluated on the Llama3-8B model, which is fine-tuned on the open-ended MedMCQA dataset.}

\label{tab:med-ood}
\end{table*}
Table~\ref{tab:med-ood} presents the experimental results of Llama3-8B on another open-ended dataset, MedMCQA. Across all OOD test sets, CogCalib demonstrates superior calibration performance, achieving the ECE compared to all baseline methods. While Deep Ensemble maintains a slight lead in ACC, CogCalib achieves comparable accuracy metrics with other baseline approaches. These findings suggest that CogCalib exhibits universal applicability across open-ended datasets.
\subsection{Results of Llama2-13B}
Table~\ref{tab:llama2-id} and Table~\ref{tab:llama2-idood} demonstrate the experimental results of CogCalib on the Llama2-13B model. For this model size, CogCalib achieves significant improvements in calibration performance while maintaining baseline accuracy. Furthermore, it demonstrates superior accuracy across multiple datasets. These experimental findings validate that CogCalib maintains its robustness when applied to larger-scale models, effectively preserving fine-tuning performance while enhancing calibration under both ID and distribution shift scenarios.
\begin{table*}
\centering
\small
\setlength{\tabcolsep}{4pt} 
\begin{tabular}{cccccc|ccc} 
		\toprule 
		\textbf{Dataset} & \textbf{Metric} & \textbf{Vanilla SFT} & \textbf{MCD} & \textbf{Ensemble} & \textbf{TS} & \textbf{CoLS} & \textbf{CoMbLS} & \textbf{CoECP} \\
		\midrule
		\multirow{2}{*}{\textbf{OBQA}} 
        & \textbf{ACC$\uparrow$} & 73.60 & 73.40 & 77.20 & 73.60 & 76.20 & \textbf{78.60} & 77.40 \\
		& \textbf{ECE$\downarrow$} & 20.91 & 16.70 & 10.10 & 18.90 & 7.55 & 6.10 & \textbf{2.62} \\
		
		\midrule
		\multirow{2}{*}{\textbf{ARC-C}} 
        & \textbf{ACC$\uparrow$} & 70.90 & 70.73 & \textbf{72.10} & 70.90 & 71.16 & 71.50 & 70.56 \\
		& \textbf{ECE$\downarrow$} & 25.98 & 22.14 & 18.21 & 24.70 & \textbf{13.34} & 13.36 & 14.43 \\
		
		\midrule
		\multirow{2}{*}{\textbf{WG-S}} 
        & \textbf{ACC$\uparrow$} & 74.59 & 74.35 & \textbf{74.98} & 74.59 & 73.56 & 72.45 & 72.63 \\
		& \textbf{ECE$\downarrow$} & 16.96 & 15.73 & 15.95 & 14.90 & \textbf{7.99} & 8.34 & 16.00 \\
		
		\midrule
		\multirow{2}{*}{\textbf{WG-M}} 
        & \textbf{ACC$\uparrow$} & 82.08 & 81.61 & \textbf{82.40} & 82.08 & 80.82 & 81.22 & 81.06 \\
		& \textbf{ECE$\downarrow$} & 16.63 & 14.58 & 12.43 & 15.60 & 7.44 & 7.17 & \textbf{6.71} \\
		
		\midrule
		\multirow{2}{*}{\textbf{BoolQ}} 
        & \textbf{ACC$\uparrow$} & 89.85 & 89.94 & 89.85 & 89.85 & 89.72 & \textbf{90.06} & 89.17 \\
		& \textbf{ECE$\downarrow$} & 9.59 & 8.69 & 7.58 & 9.40 & \textbf{2.04} & 2.47 & 4.49 \\
		
		\bottomrule
\end{tabular}
\caption{Comparison of our method's performance against baselines on in-domain (ID) datasets. Results are evaluated on Llama2-13B model fine-tuned by LoRA on 5 widely used domain-specific datasets.}
\label{tab:llama2-id}
\end{table*}
\begin{table*}
\centering
\small
\begin{tabular}{cc|c|cc|cccc} 
		\toprule 
		\multirow{2}{*}{\textbf{Metric}} &\multirow{2}{*}{\textbf{Methods}} &\multicolumn{1}{c}{\textbf{ID}} &  \multicolumn{2}{|c|}{\textbf{Smaller Distribution Shift}}&
		\multicolumn{4}{|c}{\textbf{Larger Distribution Shift}}
		\\& & OBQA &ARC-C  & ARC-E & Business & Culture & History & Psychology \\
		\midrule
\multirow{8}{*}{\textbf{ECE}$\downarrow$}&
\textbf{Vanilla SFT}& 20.91 & 26.93 & 21.12 & 20.86 & 22.41 & 22.41 & 30.40 \\
& \textbf{MCD} & 16.70 & 23.78 & 18.07 & 17.50 & 20.26 & 15.41 & 27.22 \\
& \textbf{Ensemble} & 10.10 & 18.87 & 14.27 & 17.76 & 18.26 & 14.07 & 25.80 \\
& \textbf{TS} & 18.90 & 25.00 & 19.70 & 20.90 & 21.00 & 22.70 & 28.80 \\
\cmidrule{2-9} 
& \textbf{CoLS} & 7.55 & 13.48 & 8.51 & 11.65 & \textbf{13.73} & 12.45 & 20.78 \\	
& \textbf{CoMbLS} & 6.10 & 12.10 & 7.00 & \textbf{11.42} & 13.90 & 11.72 & 19.98 \\	
& \textbf{CoECP} & \textbf{2.62} & \textbf{8.60} & \textbf{3.30} & 12.49 & 14.63 & \textbf{9.77} & \textbf{19.78} \\
\cmidrule{1-9} 

\multirow{8}{*}{\textbf{ACC}$\uparrow$}&
\textbf{Vanilla SFT}& 73.60 & 67.32 & 74.37 & 75.06 & 72.89 & 69.89 & 63.61 \\
& \textbf{MCD} & 73.40 & 66.72 & 74.03 & \textbf{75.51} & 72.89 & 68.49 & 63.70 \\
& \textbf{Ensemble} & 77.20 & 68.86 & 76.47 & \textbf{75.51} & 73.19 & 67.31 & 64.30 \\
& \textbf{TS} & 73.60 & 67.32 & 74.37 & 75.06 & 72.89 & 69.89 & 63.61 \\
\cmidrule{2-9} 
& \textbf{CoLS} & 76.20 & 68.94 & 76.18 & \textbf{75.51} & \textbf{74.10} & 70.97 & 64.39 \\	
& \textbf{CoMbLS} & 78.60 & \textbf{69.20} & \textbf{76.98} & \textbf{75.51} & 73.19 & \textbf{71.61} & 64.30 \\	
& \textbf{CoECP} & \textbf{77.40} & 67.83 & 75.34 & 75.29 & 72.89 & 71.08 & \textbf{65.25} \\

\bottomrule
\end{tabular}
\caption{Comparison of our method's performance against baselines on distribution shift datasets is presented. Results are evaluated on Llama2-13B model which is fine-tuned on the OBQA dataset.}
\label{tab:llama2-idood}
\end{table*}
\subsection{Results of Qwen2.5-7B}
We conducted comprehensive evaluations on the Qwen2.5-7B model, with results presented in Table~\ref{tab:qwen2.5-id} and Table~\ref{tab:qwen2.5-idood}. Although Qwen demonstrated superior accuracy across all datasets and inherently low ECE compared to other LLMs, our CogCalib framework still achieved significant calibration improvements over the baseline. In the in-distribution (ID) testing (Table~\ref{tab:qwen2.5-id}), both CoLS and CoMbLS consistently outperformed other approaches, while maintaining accuracy comparable to the best-performing Ensemble methods. For out-of-distribution (OOD) scenarios (Table~\ref{tab:qwen2.5-idood}), CoLS and CoMbLS achieved optimal performance across all distribution shift conditions. Notably, CoECP exhibited competitive accuracy performance under multiple larger distribution shift scenarios.

\begin{table*}
\centering
\small
\setlength{\tabcolsep}{4pt} 
\begin{tabular}{cccccc|ccc} 
		\toprule 
		\textbf{Dataset} & \textbf{Metric} & \textbf{Vanilla SFT} & \textbf{MCD} & \textbf{Ensemble} & \textbf{TS} & \textbf{CoLS} & \textbf{CoMbLS} & \textbf{CoECP} \\
		\midrule
		\multirow{2}{*}{\textbf{OBQA}} 
        & \textbf{ACC$\uparrow$} & 90.60 & 90.80 & \textbf{91.80} & 91.60 & 91.60 & \textbf{91.80} & 91.20 \\
		& \textbf{ECE$\downarrow$} & 8.48 & 8.09 & 5.05 & 7.50 & 5.65 & \textbf{3.77} & 7.51 \\
		
		\midrule
		\multirow{2}{*}{\textbf{ARC-C}} 
        & \textbf{ACC$\uparrow$} & 87.46 & \textbf{88.57} & 87.54 & 88.48 & 87.63 & 87.97 & 88.48 \\
		& \textbf{ECE$\downarrow$} & 11.41 & 8.89 & 9.76 & 9.90 & 4.24 & \textbf{3.22} & 9.37 \\
		
		\midrule
		\multirow{2}{*}{\textbf{WG-S}} 
        & \textbf{ACC$\uparrow$} & 78.37 & 78.93 & \textbf{79.87} & 79.40 & 76.95 & 78.37 & 78.45 \\
		& \textbf{ECE$\downarrow$} & 19.50 & 14.84 & 14.89 & 14.40 & 11.06 & \textbf{10.41} & 13.64 \\
		
		\midrule
		\multirow{2}{*}{\textbf{WG-M}} 
        & \textbf{ACC$\uparrow$} & 83.90 & 83.50 & \textbf{85.16} & 83.90 & 84.53 & 83.74 & 84.61 \\
		& \textbf{ECE$\downarrow$} & 15.17 & 11.44 & 10.21 & 14.70 & \textbf{4.20} & 4.84 & 6.77 \\
		
		\midrule
		\multirow{2}{*}{\textbf{BoolQ}} 
        & \textbf{ACC$\uparrow$} & 89.72 & 90.28 & \textbf{90.61} & 90.21 & 90.37 & 90.09 & 89.91 \\
		& \textbf{ECE$\downarrow$} & 9.71 & 7.91 & 7.10 & 8.80 & \textbf{1.45} & 4.22 & 7.19 \\
		
		\bottomrule
\end{tabular}
\caption{Comparison of our method's performance against baselines on in-domain (ID) datasets. Results are evaluated on Qwen2.5-7B model fine-tuned by LoRA on 5 widely used domain-specific datasets.}
\label{tab:qwen2.5-id}
\end{table*}
\begin{table*}
\centering
\small
\begin{tabular}{cc|c|cc|cccc}
\toprule 
\multirow{2}{*}{\textbf{Metric}} & \multirow{2}{*}{\textbf{Methods}} & \multicolumn{1}{c}{\textbf{ID}} & \multicolumn{2}{|c|}{\textbf{Smaller Distribution Shift}} & \multicolumn{4}{c}{\textbf{Larger Distribution Shift}} \\
& & OBQA & ARC-C & ARC-E & Business & Culture & History & Psychology \\
\midrule
\multirow{10}{*}{\textbf{ACC}$\uparrow$} & \textbf{Vanilla SFT} & 90.60 & 87.54 & 90.82 & 88.10 & \textbf{85.24} & 85.48 & 83.32 \\
& \textbf{MCD} & 90.80 & 87.12 & 90.49 & 86.96 & 84.04 & 84.85 & 82.89 \\
& \textbf{Ensemble} & \textbf{91.80} & \textbf{88.05} & \textbf{91.04} & \textbf{88.79} & 83.43 & 85.48 & 83.75 \\
& \textbf{TS} & 91.60 & 87.54 & 90.82 & 88.10 & \textbf{85.24} & 85.48 & 83.32 \\
\cmidrule{2-9}
& \textbf{CoLS} & 91.60 & 87.63 & 90.15 & 87.41 & 83.73 & 85.70 & 82.54 \\
& \textbf{CoMbLS} & \textbf{91.80} & 87.63 & 90.91 & 88.10 & 81.63 & 85.27 & 83.23 \\
& \textbf{CoECP} & 91.20 & 87.29 & 90.99 & 87.41 & \textbf{85.24} & \textbf{85.91} & \textbf{83.75} \\
\midrule
\multirow{7}{*}{\textbf{ECE}$\downarrow$} & \textbf{Vanilla SFT} & 8.48 & 11.51 & 8.60 & 10.78 & 13.66 & 12.65 & 15.09 \\
& \textbf{MCD} & 8.09 & 7.54 & 8.29 & 9.09 & 9.90 & 9.00 & 11.63 \\
& \textbf{Ensemble} & 5.05 & 9.27 & 7.38 & 8.51 & 12.17 & 10.51 & 13.08 \\
& \textbf{TS} & 7.50 & 11.20 & 8.30 & 9.90 & 13.00 & 12.10 & 14.60 \\
\cmidrule{2-9}
& \textbf{CoLS} & 5.05 & \textbf{0.75} & 2.43 & \textbf{6.34} & \textbf{5.95} & \textbf{4.80} & \textbf{5.10} \\
& \textbf{CoMbLS} & \textbf{3.77} & 2.56 & \textbf{1.99} & \textbf{6.34} & 8.17 & 4.90 & 6.90 \\
& \textbf{CoECP} & 7.51 & 8.36 & 8.33 & 7.66 & 6.18 & 6.75 & 6.66 \\
\bottomrule
\end{tabular}
\caption{Comparison of our method's performance against baselines on distribution shift datasets is presented. Results are evaluated on Qwen2.5-7B model which is fine-tuned on the OBQA dataset.}
\label{tab:qwen2.5-idood}
\end{table*}

\subsection{Results of Mistral-7B-v0.3}
We evaluated CogCalib's performance on Mistral-7B-v0.3, with results presented in Table~\ref{tab:mistral-id} and Table~\ref{tab:mistral-idood}. For in-domain testing (Table~\ref{tab:mistral-id}), our approach demonstrated superior calibration metrics compared to all baselines. While the Ensemble method achieved optimal accuracy in most cases, our method maintained competitive accuracy scores. In out-of-distribution (OOD) scenarios (Table~\ref{tab:mistral-idood}), our approach outperformed the baselines in both ECE and accuracy metrics.
\begin{table*}
\centering
\small
\setlength{\tabcolsep}{4pt} 
\begin{tabular}{cccccc|ccc} 
		\toprule 
		\textbf{Dataset} & \textbf{Metric} & \textbf{Vanilla SFT} & \textbf{MCD} & \textbf{Ensemble} & \textbf{TS} & \textbf{CoLS} & \textbf{CoMbLS} & \textbf{CoECP} \\
		\midrule
		\multirow{2}{*}{\textbf{OBQA}} 
        & \textbf{ACC$\uparrow$} & 68.40 & 67.60 & 65.40 & 68.40 & \textbf{72.60} & 70.20 & 69.00 \\
		& \textbf{ECE$\downarrow$} & 25.85 & 22.65 & 18.01 & 23.70 & \textbf{9.61} & 12.70 & 14.43 \\
		
		\midrule
		\multirow{2}{*}{\textbf{ARC-C}} 
        & \textbf{ACC$\uparrow$} & 77.13 & 77.47 & \textbf{79.18} & 77.13 & 78.24 & 77.47 & 77.30 \\
		& \textbf{ECE$\downarrow$} & 20.99 & 17.86 & 13.16 & 20.10 & \textbf{8.22} & 9.39 & 8.90 \\
		
		\midrule
		\multirow{2}{*}{\textbf{WG-S}} 
        & \textbf{ACC$\uparrow$} & 77.11 & 77.43 & \textbf{80.19} & 77.11 & 78.06 & 78.14 & 77.74 \\
		& \textbf{ECE$\downarrow$} & 21.34 & 19.41 & 13.36 & 20.40 & 11.17 & \textbf{9.48} & 10.84 \\
		
		\midrule
		\multirow{2}{*}{\textbf{WG-M}} 
        & \textbf{ACC$\uparrow$} & 83.11 & 83.27 & \textbf{83.90} & 83.11 & 82.95 & 83.35 & 82.24 \\
		& \textbf{ECE$\downarrow$} & 15.47 & 14.35 & 10.47 & 14.70 & 6.27 & \textbf{4.70} & 6.10 \\
		
		\midrule
		\multirow{2}{*}{\textbf{BoolQ}} 
        & \textbf{ACC$\uparrow$} & 89.57 & 89.60 & \textbf{90.89} & 89.57 & 89.94 & 89.48 & 90.21 \\
		& \textbf{ECE$\downarrow$} & 10.22 & 9.31 & 6.09 & 10.00 & 1.28 & \textbf{0.56} & 1.45 \\
		
		\bottomrule
\end{tabular}
\caption{Comparison of our method's performance against baselines on in-domain (ID) datasets. Results are evaluated on Mistral-7B model fine-tuned by LoRA on 5 widely used domain-specific datasets.}
\label{tab:mistral-id}
\end{table*}
\begin{table*}
\centering
\small
\begin{tabular}{cc|c|cc|cccc} 
		\toprule 
		\multirow{2}{*}{\textbf{Metric}} &\multirow{2}{*}{\textbf{Methods}} &\multicolumn{1}{c}{\textbf{ID}} &  \multicolumn{2}{|c|}{\textbf{Smaller Distribution Shift}}&
		\multicolumn{4}{c}{\textbf{Larger Distribution Shift}}
		\\& & OBQA &ARC-C  & ARC-E & Business & Culture & History & Psychology \\
		\midrule
\multirow{8}{*}{\textbf{ECE}$\downarrow$}&
\textbf{Vanilla SFT}& 20.91 & 26.93 & 21.12 & 20.86 & 22.41 & 22.41 & 30.40 \\
& \textbf{MCD} & 16.70 & 23.78 & 18.07 & 17.50 & 20.26 & 15.41 & 27.22 \\
& \textbf{Ensemble} & 10.10 & 18.87 & 14.27 & 17.76 & 18.26 & 14.07 & 25.80 \\
& \textbf{TS} & 18.90 & 25.00 & 19.70 & 20.90 & 21.00 & 22.70 & 28.80 \\
\cmidrule{2-9} 
& \textbf{CoLS} & 7.55 & 13.48 & 8.51 & 11.65 & \textbf{13.73} & 12.45 & 20.78 \\	
& \textbf{CoMbLS} & 6.10 & 12.10 & 7.00 & \textbf{11.42} & 13.90 & 11.72 & 19.98 \\	
& \textbf{CoECP} & \textbf{2.62} & \textbf{8.60} & \textbf{3.30} & 12.49 & 14.63 & \textbf{9.77} & \textbf{19.78} \\
\cmidrule{1-9} 

\multirow{8}{*}{\textbf{ACC}$\uparrow$}&
\textbf{Vanilla SFT}& 73.60 & 67.32 & 74.37 & 75.06 & 72.89 & 69.89 & 63.61 \\
& \textbf{MCD} & 73.40 & 66.72 & 74.03 & \textbf{75.51} & 72.89 & 68.49 & 63.70 \\
& \textbf{Ensemble} & 77.20 & 68.86 & 76.47 & \textbf{75.51} & 73.19 & 67.31 & 64.30 \\
& \textbf{TS} & 73.60 & 67.32 & 74.37 & 75.06 & 72.89 & 69.89 & 63.61 \\
\cmidrule{2-9} 
& \textbf{CoLS} & 76.20 & 68.94 & 76.18 & \textbf{75.51} & \textbf{74.10} & 70.97 & 64.39 \\	
& \textbf{CoMbLS} & \textbf{78.60} & \textbf{69.20} & \textbf{76.98} & \textbf{75.51} & 73.19 & \textbf{71.61} & 64.30 \\	
& \textbf{CoECP} & 77.40 & 67.83 & 75.34 & 75.29 & 72.89 & 71.08 & \textbf{65.25} \\

\bottomrule
\end{tabular}
\caption{Comparison of our method's performance against baselines on distribution shift datasets is presented. Results are evaluated on Mistral-7B model which is fine-tuned on the OBQA dataset.}
\label{tab:mistral-idood}
\end{table*}

\subsection{Results of Llama3-8B Using FFT}
\label{appendix:fft_ret}
\begin{table*}
\centering
\small
\setlength{\tabcolsep}{4pt} 
\begin{tabular}{cccccccc} 
		\toprule 
		\textbf{Dataset} & \textbf{Metric} & \textbf{Vanilla SFT} & \textbf{MCD} & \textbf{TS} & \textbf{CoLS} & \textbf{CoMbLS} & \textbf{CoECP} \\
		\midrule
		\multirow{2}{*}{\textbf{ARC-C}} 
        & \textbf{ACC$\uparrow$} & 66.81 & 65.96 & 66.81 & 70.82 & \textbf{72.18} & 70.22 \\
		& \textbf{ECE$\downarrow$} & 29.84 & 28.31 & 28.61 & 13.09 & 14.19 & \textbf{8.66} \\
		
		\midrule
		\multirow{2}{*}{\textbf{ARC-E}} 
        & \textbf{ACC$\uparrow$} & \textbf{75.04} & 74.71 & \textbf{75.04} & 72.85 & 73.99 & 74.92 \\
		& \textbf{ECE$\downarrow$} & 17.38 & 16.18 & 14.14 & 13.04 & 11.54 & \textbf{4.28} \\
		\bottomrule
\end{tabular}
\caption{Comparison of our method's performance against baselines on in-domain (ID) datasets. Results are evaluated on Llama3-8B model using full parameter fine-tuning (FFT) on 2 widely used domain-specific datasets.}
\label{tab:fft-id}
\end{table*}
\begin{table*}
\centering
\small
\begin{tabular}{cccccccc}
\toprule
\multirow{2}{*}{\textbf{Metric}} & \multirow{2}{*}{\textbf{Methods}} & \multicolumn{1}{c}{\textbf{ID}} & \multicolumn{5}{c}{\textbf{OOD}} \\
& & \textbf{ARC-C} & \textbf{ARC-E} & \textbf{OBQA} & \textbf{Business} & \textbf{History} & \textbf{Psychology} \\
\midrule
\multirow{6}{*}{\textbf{ACC}$\uparrow$} &
\textbf{Vanilla SFT} & 66.81 & 74.87 & 68.40 & 72.31 & 64.52 & 55.57 \\
& \textbf{MCD} & 65.96 & 74.71 & 69.40 & 72.31 & 63.23 & 55.14 \\
& \textbf{TS} & 66.81 & 74.87 & 68.40 & 72.31 & 64.52 & 55.57 \\
& \textbf{CoLS} & 70.82 & 72.85 & \textbf{73.80} & 71.85 & 70.97 & \textbf{67.42} \\
& \textbf{CoMbLS} & \textbf{72.18} & 73.99 & 67.60 & 72.31 & \textbf{72.37} & 64.82 \\
& \textbf{CoECP} & 70.22 & \textbf{74.92} & 65.60 & \textbf{75.74} & 69.14 & 66.64 \\
\midrule
\multirow{6}{*}{\textbf{ECE}$\downarrow$} &
\textbf{Vanilla SFT} & 29.84 & 22.63 & 27.99 & 22.29 & 30.29 & 40.19 \\
& \textbf{MCD} & 28.31 & 20.87 & 23.59 & 20.10 & 28.67 & 37.72 \\
& \textbf{TS} & 28.61 & 21.60 & 26.20 & 20.90 & 28.60 & 38.40 \\
& \textbf{CoLS} & 13.09 & 13.04 & \textbf{8.00} & 9.60 & 12.73 & 16.14 \\
& \textbf{CoMbLS} & 14.19 & 11.54 & 15.72 & 11.45 & 13.32 & 20.69 \\
& \textbf{CoECP} & \textbf{8.66} & \textbf{4.28} & 9.47 & \textbf{7.90} & \textbf{10.66} & \textbf{11.07} \\
\bottomrule
\end{tabular}
\caption{Comparison of our method's performance against baselines on OOD datasets. Results are evaluated on Llama3-8B model using full parameter fine-tuning (FFT) on OOD scenarios.}
\label{tab:fft-ood}
\end{table*}
In addition to LoRA, we validated CogCalib using Full-parameter Fine-Tuning (FFT) on the Llama3-8B model. Table~\ref{tab:fft-id} and Table~\ref{tab:fft-ood} present the results for ID and OOD evaluations, respectively, demonstrating that CogCalib is effectively applicable to FFT. The method not only significantly enhances calibration performance but also improves model generalization. Specifically, the model trained on ARC-C achieved an average accuracy of 70.67\% across other tasks, surpassing the conventional SFT baseline (67.13\%). We hypothesize that the introduction of the calibration term mitigates overfitting during fine-tuning, thereby enhancing the model's generalization capabilities.

\begin{figure*}[t]
    \centering
    \includegraphics[width=\linewidth]{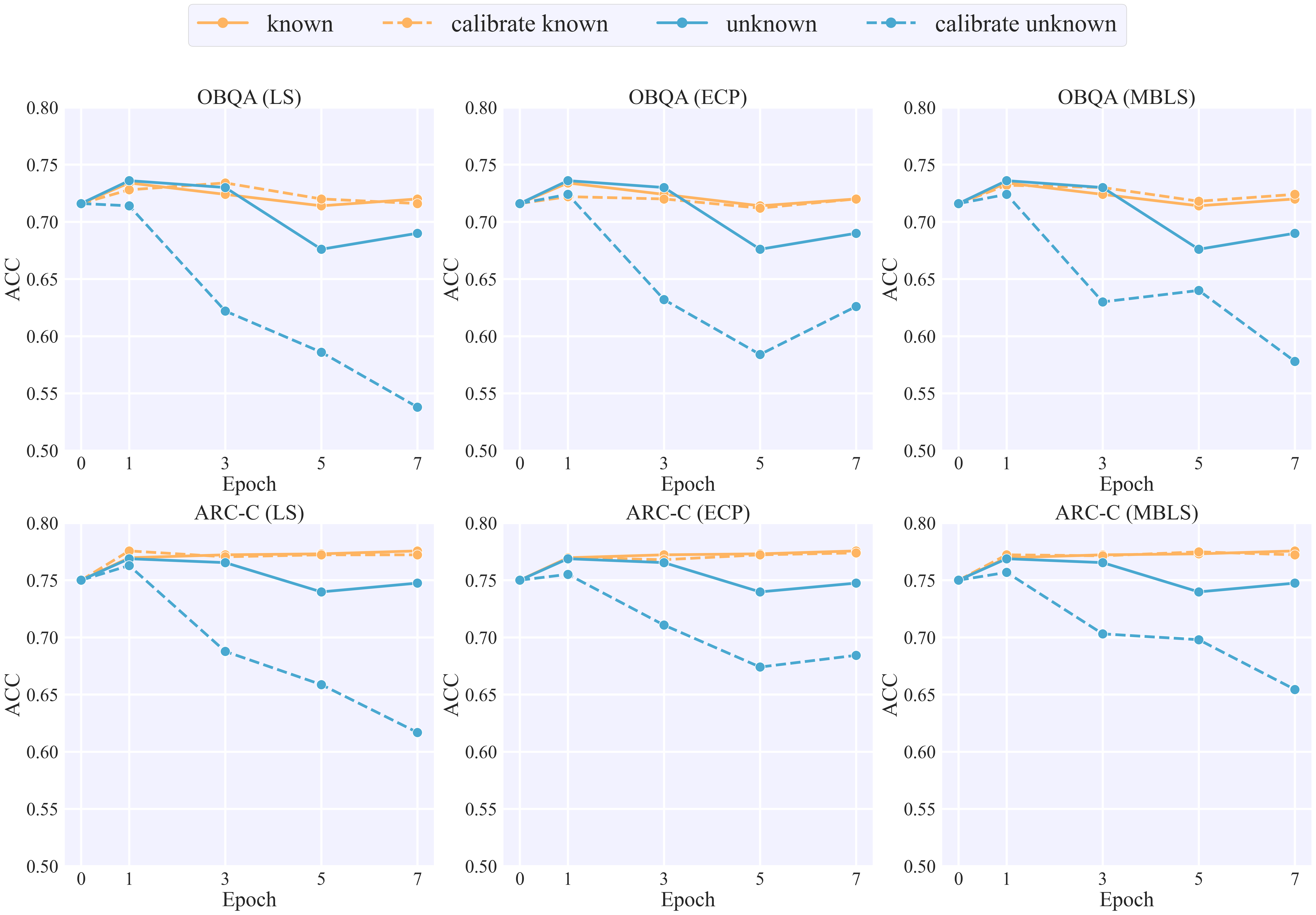}
    \caption{Comparison of accuracy on downstream tasks with and without calibration methods when fine-tuning on MedMCQA known or unknown data. It is observed that calibrating on unknown data significantly deteriorates the performance of other out-of-distribution (OOD) tasks.}
    \label{fig:cal_unk_impact}
\end{figure*}

\subsection{Effects of Calibrating Unknown Data}
\label{appendix:effects}
We investigated the impact of calibrating unknown data on the model's downstream task performance. Beyond the adverse effects on OOD tasks observed with multiple-choice data in Section~\ref{subsec:abala}, we found that this negative impact was even more pronounced when using open-ended data as the fine-tuning dataset. 

In our experimental setup, we utilized equal amounts (1k samples) of known or unknown data from MedMCQA as fine-tuning datasets, implementing various calibration enhancement methods. As illustrated in the Figure~\ref{fig:cal_unk_impact}, while using calibration methods on known data did not significantly affect performance on OOD tasks, applying calibration methods to unknown data led to a consistent decline in OOD performance. These findings underscore the critical importance of thoroughly learning from unknown data during the fine-tuning process.

\subsection{Comparision to Vanilla and Random Calibration.}
\label{appendix:vanilla_vs_random}
\begin{figure}
    \centering
    \includegraphics[width=\linewidth]{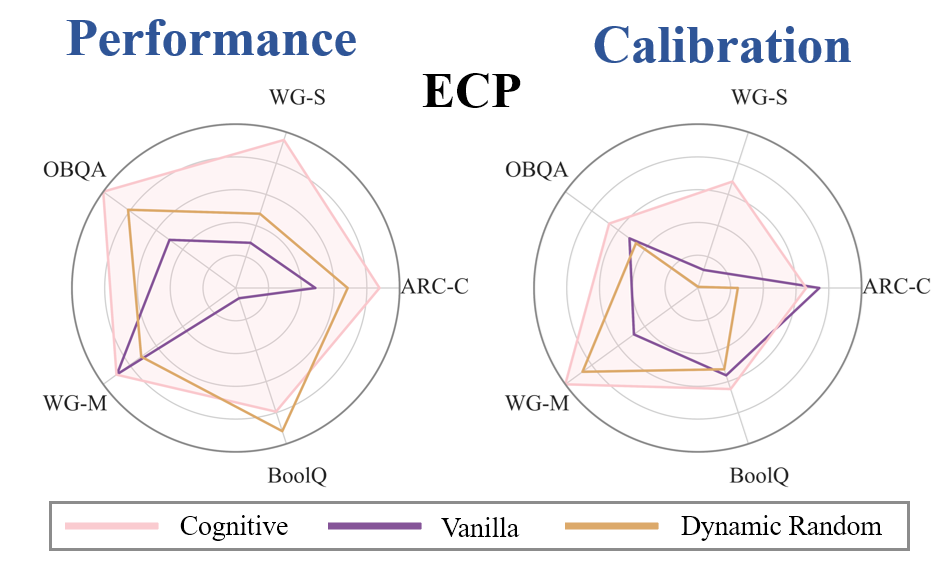}
    \caption{Comparsion between CogECP, vanilla ECP, and dynamic random ECP. Our proposed method, CoECP, consistently outperforms all baseline methods in terms of both calibration and accuracy across multiple datasets.}
    \label{fig:ablation_1_ecp}
\end{figure}

We present a comparative analysis of CoECP against its corresponding variants: Vanilla calibration and Random Calibration. As illustrated in Figure~\ref{fig:ablation_1_ecp}, CoECP consistently outperforms both baseline methods across most datasets in terms of calibration and fine-tuning performance. These results further support our findings in Section~\ref{sec:3}, which emphasize that unknown data plays a crucial role in aligning the model with downstream tasks. The effective utilization of such data simultaneously enhances fine-tuning performance and improves calibration metrics.
\subsection{Sensitivity to Threshold Computation.}
\label{appendix:sense_to_threshold}
We compare our threshold calculation called \textbf{Balanced}, with the \textbf{Accuracy-based} method seeking threshold $t$ which achieves best known/unknown data classification accuracy using negative log-likelihood.

\begin{figure}[t] 
    \centering
    \includegraphics[width=\linewidth]{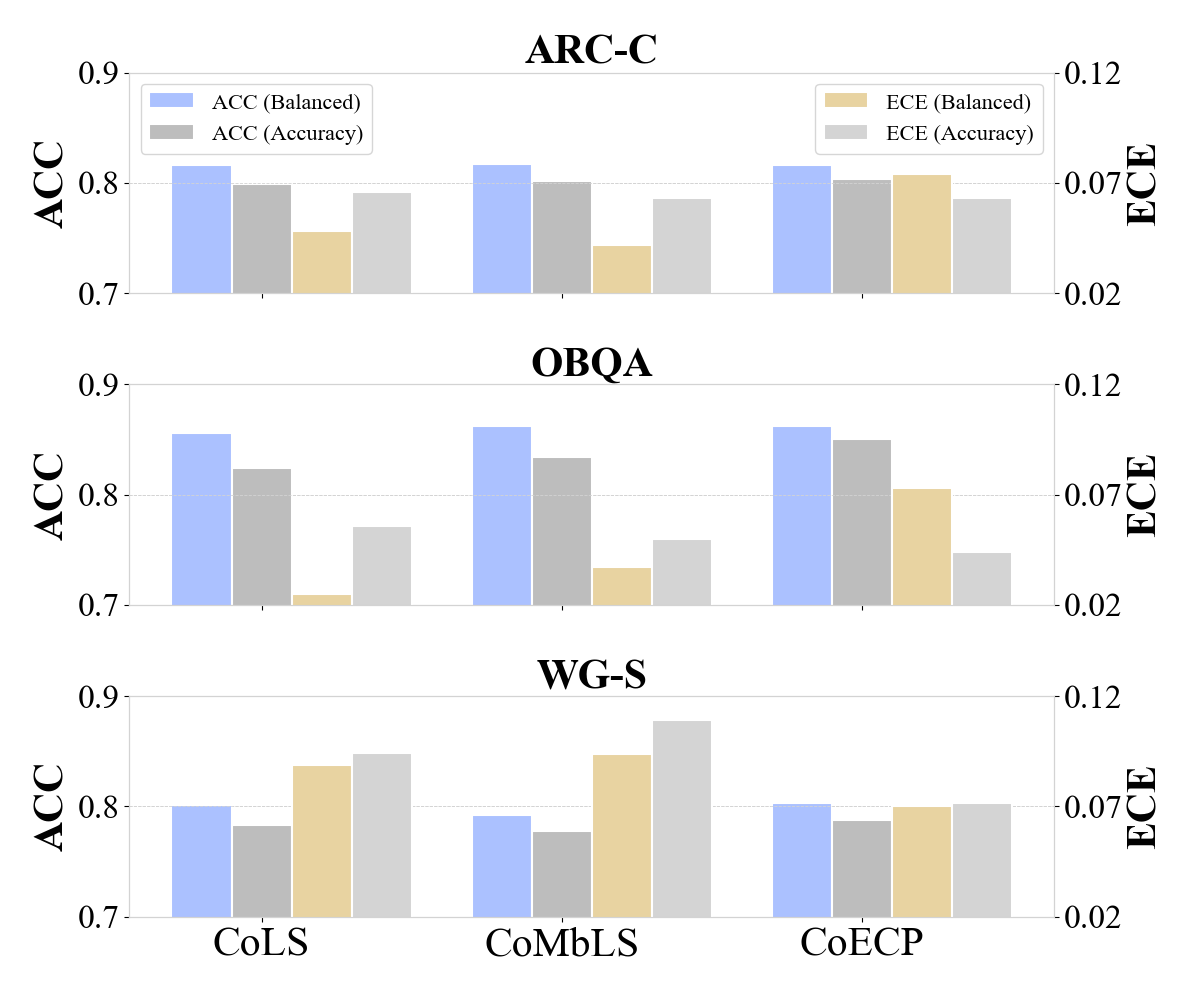}
    \caption{Sensitivity to Threshold Computation. Our method demonstrates robust performance across different threshold calculation approaches, while employing our proposed threshold computation methodology yields superior fine-tuning performance.}
    \label{fig:adjust}
\end{figure}

Figure~\ref{fig:adjust} reveals that these 2 calculations have similar calibration effects, but our method attains a higher accuracy. This improvement can be attributed to the higher TNR achieved by our approach (Table~\ref{tab:performance_threshhold}). Accuracy-based calculation tends to misclassify unknown samples, consequently applying calibration to these samples as well, which prevents LLM from effectively learning critical knowledge.

In detail, Table~\ref{tab:performance_threshhold} presents the classification results for both known and unknown data during the fine-tuning under different threshold calculation methods. The results demonstrate that when using our method, both TPR and TNR are well-balanced. In contrast, when employing the highest accuracy for threshold calculation, the TNR exhibits notably lower values. This discrepancy indicates that the latter method incorrectly classifies unknown data as known data and subsequently applies calibration methods, preventing the model from effectively learning crucial patterns in unknown data during the fine-tuning process. This limitation explains the consistently inferior performance of this method compared to the former approach, as illustrated in Figure~\ref{fig:adjust}.
\begin{table*}
\centering
\small
\begin{tabular}{ccccccc}
\toprule
\multirow{2}{*}{\textbf{Dataset}} & \multicolumn{3}{c}{\textbf{argmax(TNR+TPR)}} & \multicolumn{3}{c}{\textbf{argmax(Accuracy)}} \\
\cmidrule(lr){2-4} \cmidrule(lr){5-7}
 & \textbf{Accuracy} & \textbf{TPR} & \textbf{TNR} & \textbf{Accuracy} & \textbf{TPR} & \textbf{TNR} \\
\midrule
\textbf{ARC-C} & 99.51 & 99.54 & 99.15 & 99.58 & 99.95 & 95.18 \\
\textbf{OBQA} & 99.44 & 99.44 & 99.52 & 99.78 & 99.91 & 96.82 \\
\textbf{WG-S} & 98.83 & 98.77 & 99.52 & 99.49 & 99.83 & 96.41 \\
\bottomrule
\end{tabular}
\caption{Performance metrics (Accuracy, True Positive Rate (TPR), and True Negative Rate (TNR)) for different optimization criteria on the ARC-C, OBQA, and WG-S datasets.}
\label{tab:performance_threshhold}
\end{table*}

\begin{table}[h]
\centering
\small
\begin{tabular}{cccccc}
\toprule 
\multirow{2}{*}{\textbf{Dataset}} & \multirow{2}{*}{\textbf{Metrics}} & \multicolumn{4}{c}{\textbf{ECP factor}} \\
& & \textbf{0.05} & \textbf{0.075} & \textbf{0.1} & \textbf{0.125} \\
\midrule
\multirow{2}{*}{\textbf{ARC-C}} & \textbf{ACC} & 82.20 & 81.80 & 81.50 & 80.90 \\
& \textbf{ECE} & 15.38 & 15.66 & \textbf{7.21} & 15.85 \\
\midrule
\multirow{2}{*}{\textbf{wino-S}} & \textbf{ACC} & 80.20 & 78.70 & 79.20 & 79.20  \\
& \textbf{ECE} & 17.05 & 18.72 & \textbf{2.38} & 6.28  \\
\bottomrule
\end{tabular}
\caption{Performance metrics (ACC and ECE) for different ECP factors on the ARC-C and wino-S datasets. }
\label{tab:hyper_ecp}
\end{table}

\subsection{Sensitivity to Hyperparameters.}
\label{appendix:sens_hyper}
We demonstrated additional robustness results regarding CogCalib hyperparameters in Table~\ref{tab:hyper_ecp}, Table~\ref{tab:hyper_mbls_factor_margin_0} and Table~\ref{tab:hyper_mbls_factor_margin_5}. By varying the ECP Factor from 0.05 to 0.125 and MbLS Factor from 0.05 to 0.125 (with margins of 0 and 5), our method consistently achieved improved calibration performance compared to the Temperature Scaling baseline, which showed ECE values of 12.3 on ARC-C and 15.4 on Wino-S. These results across multiple parameter settings validate the robustness of our approach. However, it is worth noting that the ECP parameters may require more fine-grained tuning for optimal performance.

\begin{table}[H]
\centering
\small
\begin{tabular}{cccccc}
\toprule 
\multirow{2}{*}{\textbf{Dataset}} & \multirow{2}{*}{\textbf{Metrics}} & \multicolumn{4}{c}{\textbf{MbLS Factor (Margin=0)}} \\
& & \textbf{0.05} & \textbf{0.075} & \textbf{0.1} & \textbf{0.125} \\
\midrule
\multirow{2}{*}{\textbf{ARC-C}} & \textbf{ACC} & 80.30 & 81.90 & 81.60 & 82.30 \\
& \textbf{ECE} & 12.55 & 8.87 & 6.22 & \textbf{2.37} \\
\midrule
\multirow{2}{*}{\textbf{WG-S}} & \textbf{ACC} & 77.30 & 78.10 & 79.60 & 78.60 \\
& \textbf{ECE} & 15.55 & 12.01 & 9.94 & \textbf{8.06} \\
\bottomrule
\end{tabular}
\caption{Performance metrics (ACC and ECE) for different MbLS factors with Margin=0 on the ARC-C and WG-S datasets.}
\label{tab:hyper_mbls_factor_margin_0}
\end{table}
\begin{table}[H]
\centering
\small
\begin{tabular}{cccccc}
\toprule 
\multirow{2}{*}{\textbf{Dataset}} & \multirow{2}{*}{\textbf{Metrics}} & \multicolumn{4}{c}{\textbf{MbLS Factor (Margin=5)}} \\
& & \textbf{0.05} & \textbf{0.075} & \textbf{0.1} & \textbf{0.125} \\
\midrule
\multirow{2}{*}{\textbf{ARC-C}} & \textbf{ACC} & 81.20 & 81.10 & 81.40 & 82.20 \\
& \textbf{ECE} & 11.35 & 9.63 & 5.86 & \textbf{2.53} \\
\midrule
\multirow{2}{*}{\textbf{WG-S}} & \textbf{ACC} & 78.10 & 78.20 & 78.80 & 79.30 \\
& \textbf{ECE} & 15.09 & 11.87 & 11.04 & \textbf{6.19} \\
\bottomrule
\end{tabular}
\caption{Performance metrics (ACC and ECE) for different MbLS factors with Margin=5 on the ARC-C and WG-S datasets.}
\label{tab:hyper_mbls_factor_margin_5}
\end{table}

\section{Implementation Details}
In this section, we present a detailed analysis of the SliCK method, the implementation of Temperature Scaling for both open-ended and multiple-choice data, along with our specific hyperparameter configurations.
\subsection{Details of SliCK}
\label{appendix:SliCK}
In Section~\ref{sec:3}, we employed the SliCK~\citep{gekhman2024does} to classify known and unknown data. Specifically, the SliCK method concatenates 10 different randomly selected 4-shot prompts for each question-answer pair and performs 16 sampling iterations with temperature settings of either 0 or 0.5. 

The prediction accuracy under greedy decoding is denoted as $P(T = 0)$, while $P(T > 0)$ represents the prediction accuracy when $T = 0.5$. Based on the accuracy calculations from multiple sampling iterations, the data is categorized into 4 classes: HighlyKnown, MaybeKnown, WeaklyKnown, and Unknown as shown in Table~\ref{tab:type_of_data}. For the experiments conducted in Section~\ref{sec:3}, we treated HighlyKnown samples as known data and maintained the Unknown classification as is.
\begin{table}[H]
\centering
\small
\begin{tabular}{cc}
\toprule
\textbf{Type} & \textbf{Definition}\\
\midrule
HighlyKnown & $P(T=0)=1$  \\
MaybeKnown & $P(T=0) \in (0,1)$   \\
WeaklyKnown & $P(T=0)=0 \wedge P(T>0)>0$   \\
Unknown & $P(T \geq 0) =0 $  \\
\bottomrule
\end{tabular}
\caption{SliCK's definition of different type of data.}
\label{tab:type_of_data}
\end{table}

\subsection{Details of Temperature Scaling}
\label{appendix:ts}
For multiple-choice datasets, we employ the maximum probability of the first output token as the confidence. In contrast, for long-text datasets, we adopt the geometric mean probability as the confidence for long-text generation, following the approach proposed in LITCAB~\citep{liu2023litcab}, as illustrated in the Equation (\ref{eq:gm_conf}),
\begin{equation}
\label{eq:gm_conf}
p(y|x) = \sqrt[L]{\prod \limits_{t=1}^L p(y_t|x,y_{<t})}.
\end{equation}

To determine the optimal temperature for individual tokens, we implement the method developed by Guo et al.\footnote{https://github.com/gpleiss/temperature\_scaling} , which utilizes gradient descent to minimize the ECE loss and identify the optimal temperature parameter. For long-text confidence, we employ a grid search strategy to determine the optimal temperature. The resulting optimal temperature is then applied to enhance calibration for both ID and OOD samples, ensuring a fair comparison.

\begin{table}[h]
\centering
\small
\begin{tabular}{c|c}
\toprule 
\textbf{Hyperparamter} & \textbf{value} \\
\midrule
\textbf{LS $\epsilon$} & 0.1 \\
\textbf{MbLS $\gamma$} & 0.1 \\
\textbf{MbLS Margin} & 0 \\
\textbf{ECP $\beta$} & 0.1 \\
\bottomrule
\end{tabular}
\caption{Calibration term's hyperparameters for multi-choice QA task.}
\label{tab:calibration_term_hyper_mc_task}
\end{table}

It is important to note that the optimal temperature discovered on the validation set typically fails to improve OOD calibration, which represents a significant limitation of Temperature Scaling. This limitation becomes particularly critical in the context of LLMs, which are required to handle diverse tasks effectively.
\subsection{Hyperparameters}
\label{appendix:hyper}
In CogCalib‘s framework, we experiment with 3 calibration losses: Label Smoothing (LS), Margin-based Label Smoothing (MbLS), and ECP. Considering the distinct nature of tasks we experimented on: multiple-choice QA with concentrated probability distributions and open-ended QA with inherently higher uncertainty in outputs, we adopted 2 sets of calibration loss hyperparameters. Specifically, the hyperparameter settings for multi-choice QA and open-ended QA are presented in Table ~\ref{tab:calibration_term_hyper_mc_task} and Table ~\ref{tab:calibration_term_hyper_oe_task}, respectively. All experiments repeat three times, and the average results are recorded. Models smaller than 13B parameters are trained on an NVIDIA RTX-4090 GPU, while the 13B model is trained on an NVIDIA A100 GPU.

\begin{table}[h]
\centering
\small
\begin{tabular}{c|c}
\toprule 
\textbf{Hyperparamter} & \textbf{value} \\
\midrule
\textbf{LS $\epsilon$} & 0.15 \\
\textbf{MbLS $\gamma$} & 0.15 \\
\textbf{MbLS Margin} & 10 \\
\textbf{ECP $\beta$} & 0.15 \\
\bottomrule
\end{tabular}
\caption{Calibration term's hyperparameters for open-ended QA task.}
\label{tab:calibration_term_hyper_oe_task}
\end{table}

In our experimental setup, we fine-tune the LLM using both LoRA and FFT approaches. For LoRA implementation, we incorporate LoRA adapters into all linear layers of the LLM, maintaining the default PEFT configurations from Huggingface, as detailed in Table ~\ref{tab:lora_hyper}. The FFT parameters are specified in Table ~\ref{tab:fft_hyper}.

\begin{table}[H]
\centering
\small
\begin{tabular}{c|c}
\toprule 
\textbf{Hyperparamter} & \textbf{value} \\
\midrule
\textbf{LoRA $r$} & 8 \\
\textbf{LoRA $\alpha$} & 16 \\
\textbf{LoRA target} & all \\
\textbf{Learning Rate} & $6.0\times10^{-5}$ \\
\textbf{Batch size} & 2 \\
\textbf{Learning Rate scheduler} & Linear \\
\textbf{Max Sequence Length} & 1024 \\
\bottomrule
\end{tabular}
\caption{Experimental hyperparameters used for LoRA fine-tuning.}
\label{tab:lora_hyper}
\end{table}

\begin{table}[H]
\centering
\small
\begin{tabular}{c|c}
\toprule 
\textbf{Hyperparamter} & \textbf{value} \\
\midrule
\textbf{Learning Rate} & $1.0\times10^{-5}$ \\
\textbf{Batch size} & 4 \\
\textbf{Learning Rate scheduler} & Cosine \\
\textbf{Warmup Ratio} & 0.1 \\
\textbf{Max Sequence Length} & 1024 \\
\bottomrule
\end{tabular}
\caption{Experimental hyperparameters used for FFT fine-tuning.}
\label{tab:fft_hyper}
\end{table}

\end{document}